\documentclass[letterpaper]{article} %
\usepackage{aaai23}  %
\usepackage{times}  %
\usepackage{helvet}  %
\usepackage{courier}  %
\usepackage[hyphens]{url}  %
\usepackage{graphicx} %
\usepackage{natbib}  %
\usepackage{caption} %
\usepackage{algorithm}
\usepackage{algorithmic}

\usepackage{newfloat}
\usepackage{listings}
\DeclareCaptionStyle{ruled}{labelfont=normalfont,labelsep=colon,strut=off} %
\floatstyle{ruled}
\newfloat{listing}{tb}{lst}{}
\floatname{listing}{Listing}
\usepackage{amssymb,amsmath,algorithm,algorithmic}

\usepackage{booktabs}       %
\usepackage{amsfonts}       %
\usepackage{nicefrac}       %
\usepackage{microtype}      %
\usepackage{microtype}
\usepackage{graphicx}
\usepackage{subcaption}
\usepackage[export]{adjustbox}
\usepackage{booktabs} %
\usepackage{physics}
\usepackage{cuted}
\usepackage{multirow}
\newcommand{\ind}{\mathbf{1}}
\newcommand{\eps}{\epsilon}
\newcommand{\hz}{\hat{\rho}}
\newcommand{\hy}{\hat{r}}
\newcommand{\Var}{\textnormal{Var}}
\newcommand{\Cov}{\textnormal{Cov}}
\newcommand{\hmu}{\hat{\mu}}
\newcommand{\hmuht}{\hmu_{\textsc{HT}}}
\newcommand{\E}{\mathbb{E}}
\newcommand{\sel}{\mathcal{K}} %
\newcommand{\arms}{\mathcal{A}} %
\newcommand{\htheta}{\hat{\theta}} %
\usepackage{pifont}%
\newcommand{\xmark}{\ding{55}}%
\usepackage{changepage}

\input{defs.tex}
\newif\ifnta

\ntafalse

\newif\ifarxiv

\arxivtrue

\ifarxiv

\usepackage[dvipsnames]{xcolor}
\usepackage[colorlinks]{hyperref}
\usepackage{xpatch}
\definecolor{crimsonred}{rgb}{0.6, 0.0, 0.0}
\hypersetup{
linkcolor=crimsonred
,citecolor=crimsonred
,filecolor=Mulberry
,urlcolor=crimsonred
,menucolor=BrickRed
,runcolor=Mulberry
,linkbordercolor=BrickRed
,citebordercolor=BrickRed
,filebordercolor=Mulberry
,urlbordercolor=NavyBlue
,menubordercolor=BrickRed
,runbordercolor=Mulberry
}
\else
\usepackage{xcolor}         %

\fi

\title{Integrating Reward Maximization and Population Estimation: Sequential Decision-Making for Internal Revenue Service Audit Selection}
\author {
    Peter Henderson,\textsuperscript{\rm 1} Ben Chugg,\textsuperscript{\rm 2} Brandon Anderson,\textsuperscript{\rm 3} Kristen Altenburger,\textsuperscript{\rm 1}\\\textbf{Alex Turk,\textsuperscript{\rm 3} John Guyton,\textsuperscript{\rm 3} Jacob Goldin,\textsuperscript{\rm 4} Daniel E. Ho\textsuperscript{\rm 1}} \\
}
\affiliations {
    \textsuperscript{\rm 1} Stanford University, 
    \textsuperscript{\rm 2} Carnegie Mellon University
    \textsuperscript{\rm 3} Internal Revenue Service,
    \textsuperscript{\rm 4} University of Chicago 
}

\begin{document}

\maketitle

\begin{abstract}
We introduce a new setting, \emph{optimize-and-estimate structured bandits}.
Here, a policy must select a batch of arms, each characterized by its own context, that would allow it to both maximize reward and maintain an accurate (ideally unbiased) population estimate of the reward.
This setting is inherent to many public and private sector applications and often requires handling delayed feedback, small data, and distribution shifts.
We demonstrate its importance on real data from the United States Internal Revenue Service (IRS).
The IRS performs yearly audits of the tax base. Two of its most important objectives are to identify suspected misreporting and to estimate the “tax gap” — the global difference between the amount paid and true amount owed.
Based on a unique collaboration with the IRS, we cast these two processes as a unified optimize-and-estimate structured bandit.
We analyze optimize-and-estimate approaches to the IRS problem and propose a novel mechanism for unbiased population estimation that achieves rewards comparable to baseline approaches.
This approach has the potential to improve audit efficacy, while maintaining policy-relevant estimates of the tax gap. This has important social consequences given that the current tax gap is estimated at nearly half a trillion dollars. 
We suggest that this problem setting is fertile ground for further research and we highlight its interesting challenges. The results of this and related research are currently being incorporated into the continual improvement of the IRS audit selection methods.
\end{abstract}

\section{Introduction}
\label{submission}

Sequential decision-making algorithms, like bandit algorithms and active learning, have been used across a number of domains: from ad targeting to clinical trial optimization~\citep{bouneffouf2019survey}.
In the public sector, these methods are not yet widely adopted, but could improve the efficiency and quality of government services if deployed with care.
\citet{henderson2021} provides a review of this potential.
Many administrative enforcement agencies in the United States (U.S.) face the challenge of allocating scarce resources for auditing regulatory non-compliance.
But these agencies must also balance additional constraints and objectives simultaneously. In particular, they must maintain an accurate estimate of population non-compliance to inform policy-making.
In this paper, we focus on the potential of unifying audit processes with these multiple objectives under a sequential decision-making framework. 
We call our setting \emph{optimize-and-estimate structured bandits}.
This framework is useful in practical settings, challenging, and has the potential to bring together methods from survey sampling, bandits, and active learning.
It poses an interesting and novel challenge for the machine learning community and can benefit many public and private sector applications (see more discussion in
\ifarxiv
Appendix~\ref{app:relevance}).
\else
Appendix C).\footnote{Appendices can be found at: \url{https://arxiv.org/abs/2204.11910}.}
\fi
It is critical to many U.S. federal agencies that are bound \emph{by law} to balance enforcement priorities with population estimates of improper payments~\citep{henderson2021,omb2018requirements,omb2021requirements}.

We highlight this framework with a case study of the Internal Revenue Service (IRS).
The IRS selects taxpayers to audit every year to detect under-reported tax liability. 
Improving audit selection could yield 10:1 returns in revenue and help fund socially beneficial programs~\citep{sarin2019shrinking}.
But the agency must also provide an accurate assessment of the tax gap (the projected amount of tax under-reporting if all taxpayers were audited). 
Currently, the IRS accomplishes this via two separate mechanisms: (1) a stratified random sample to estimate the tax gap; (2) a focused risk-selected sample of taxpayers to collect under-reported taxes.
Based on a unique multiyear collaboration with the IRS, we were provided with full micro data access to masked audit data to research how machine learning could improve audit selection. We investigate whether these separate mechanisms and objectives can be combined into one batched structured bandit algorithm, which must both maximize reward and maintain accurate population estimates. 
Ideally, if information is reused, the system can make strategic selections to balance the two objectives.
We benchmark several sampling approaches and examine the trade-offs between them with the goal of understanding the effects of using bandit algorithms in this high-impact setting.
We identify several interesting results and challenges using historical taxpayer audit data in collaboration with the IRS.

First, we introduce a novel sampling mechanism called \emph{Adaptive Bin Sampling} (ABS) which guarantees an unbiased population estimate by employing a Horvitz-Thompson (HT) approach~\cite{horvitz1952generalization}, but is comparable to other methods for cumulative reward. Its unbiasedness and comparable reward comes at the cost of additional variance, though the method provides fine-grained control of this variance-reward trade-off. 

Second, we compare this approach to $\epsilon$-greedy and optimism-based approaches, where a model-based population estimate is used. We find that model-based approaches are biased absent substantial reliance on $\epsilon$, but low in variance. Surprisingly, we find that greedy approaches perform well in terms of reward, reinforcing findings by~\citet{bietti2018contextual} and \citet{bastani2021mostly}. But we find the bias from population estimates in the greedy regime to be substantial. These biases are greatly reduced even with small amounts of random exploration, but the lack of unbiasedness guarantees make them unacceptable for many public policy settings.

Third, we show that more reward-optimal approaches tend to sample high-income earners versus low-income earners. And more reward-optimal approaches tend to audit fewer tax returns that yield no change (a reward close to 0). This reinforces the importance of reducing the amount of unnecessary exploration, which would place audit burdens on compliant taxpayers.
\ifarxiv
Appendix~\ref{app:ethics}
\else
Appendix D
\fi
details other ethical and societal considerations taken into account with this work.

Fourth, we show that model errors are heteroskedastic, resulting in more audits of high-income earners by optimism-based methods, but not yielding greater rewards.\footnote{We note that it is possible that these stem from measurement limitations in the high income space~\cite{guyton2021tax}.}

We demonstrate that combining random and focused audits into a single framework can more efficiently maximize revenue while retaining accuracy for estimating the tax gap. While additional research is needed in this new and challenging domain, this work demonstrates the promise of applying a bandit-like approach to the IRS setting, and optimize-and-estimate structured bandits more broadly. 
\ifnta
\else
The results of this and related research are currently being incorporated into the continual improvement of the IRS audit selection methods.
\fi

\section{Background}
\label{sec:background}

\textbf{Related Work.} The bandit literature is large. To fully engage with it, we provide an extended literature review in
\ifarxiv
Appendix~\ref{sec:related_work}
\else
Appendix E
\fi
, but we mention several strands of related research here. The fact that adaptively collected data leads to biased estimation (whether model-based or not) is well-known. See, e.g., \citet{nie2018adaptively,xu2013estimation,shin2021bias}. A number of works have sought to develop sampling strategies that combat bias. See, e.g.~\citet{dimakopoulou2017estimation}.
This work has been in the multi-armed bandit (MAB) or (mostly linear) contextual bandit settings. 
In the MAB setting, there has also been some work which explicitly considers the trade-off between reward and model-error. See, e.g,~\citet{liu2014trading,erraqabi2017trading}. In \ifarxiv
Appendix~\ref{sec:related_work}
\else
Appendix E
\fi
we provide a comparison against our setting, but crucially we have volatile arms which make our setting different and closer to the linear stochastic bandit work (a form of structured bandit)~\citep{abbasi2011improved,joseph2018meritocratic}. However, we require non-linearity and batched selection, as well as adding the novel estimation objective to this structured bandit setting. To our knowledge, ours is the first formulation which actively incorporates bias and variance of population estimates into a batched structured bandit problem formulation. 
Moreover, our focus is to study this problem in a real-world public sector domain, taking on the challenges proposed by \citet{wagstaff2012machine}. No work we are aware of has analyzed the IRS setting in this way.

\textbf{Institutional Background.} The IRS maintains two distinct categories of audit processes. National Research Program (NRP) audits enable population estimation of non-compliance while  Operational (Op) audits are aimed at collecting taxes from non-compliant returns. The NRP is a core measurement program for the IRS to regularly evaluate tax non-compliance \citep{gao2002new,gao2003irs}.
The NRP randomly selects, via a stratified random sample, $\sim$15k tax returns each year for research audits~\citep{internal2019federal}, although this has been decreasing in recent years and there is pressure to reduce it further \citep{marr2016irs,cbo2020trends}. These audits are used to identify new areas of noncompliance, estimate the overall tax gap, and estimate improper payments of certain tax credits. Given a recent gross tax gap estimate of \$441 billion~\citep{internal2019federal}, even minor increases in efficiency can yield large returns. 
In addition to its use for tax gap estimation, NRP serves as a training set for certain Op audit programs like the Discriminant Function (DIF) System~\citep{irm}, which is based on a modified Linear Discriminant Analysis (LDA) model~\cite{dif1976}.
DIF also incorporates other measures and policy objectives that we do not consider here. We instead focus on the stylized setting of only population estimation and reward maximization.
Tax returns that have a high likelihood of a significant adjustment, as calculated by DIF, have a higher probability of being selected for Op audits. 

It is important to highlight that Op data is not used for estimating the DIF risk model and is not used for estimating the tax gap (specifically, the individual income misreporting component of the tax gap). Though NRP audits are jointly used for population estimates of non-compliance and risk model training, the original sampling design was not optimized for both revenue maximization and estimator accuracy for tax non-compliance. Random audits have been criticized for burdening compliant taxpayers and for failing to target  areas of known non-compliance ~\citep{lawsky2008fairly}. 
The current process already somewhat represents informal sequential decision-making system. NRP strata are informed by the Op distribution, and are adjusted year-to-year. 
We posit that by formalizing the current IRS system in the form of a sequential decision-making problem, we can incorporate more methods to improve its efficiency, accuracy, and fairness.

\textbf{Data.} The data used throughout this work is from the  NRP's random sample~\citep{andreoni1998tax,johns2010distribution,internal2016federal,internal2019federal}, which we will treat as the full population of audits, since they are collected via a stratified random sample and represent the full population of taxpayers. 
The NRP sample is formed by dividing the taxpayer base into activity classes based on income and claimed tax credits, and various strata within each class. Each stratum is weighted to be representative of the national population of tax filers. Then a stratified random sample is taken across the classes.
NRP audits seek to estimate the correctness of the whole return via a close to line-by-line examination~\citep{belnap2020real}. %
This differs from Op audits, which are narrower in scope and focus on specific issues. %
Given the expensive nature of NRP audits, NRP sample sizes are relatively small ($\sim$15k/year)~\citep{guyton2018effects}.
The IRS uses these audits to estimate the tax gap and average non-compliance.\footnote{The IRS uses statistical adjustments to compensate naturally occurring variation in the depth of audit, and taxpayer misreporting that is difficult to find via auditing, and other NRP sampling limitations~\citep{guyton2020tax,internal2019federal,erard2011individual}. For the goals of this work we ignore these.}
Legal requirements for these estimates exist \citep{tas2018improper}.
The 2018 Office of Management and Budget (OMB) guidelines, for instance, state that these values should be ``statistically valid'' (unbiased estimates of the mean) and have ``$\pm 3\%$ or better margin of error at the 95\% confidence level  for  the  improper  payment  percentage  estimate'' \citep{omb2018requirements}.
Later OMB guidelines have provided more discretion to programs for developing feasible point estimates and confidence intervals (CIs) \citep{omb2021requirements}. Unbiasedness remains an IRS policy priority.

 \begin{figure*}[!htbp]
    \begin{minipage}{0.5\textwidth}
        \subcaptionbox{Risk distribution and parameterizations}{\includegraphics[width=\textwidth]{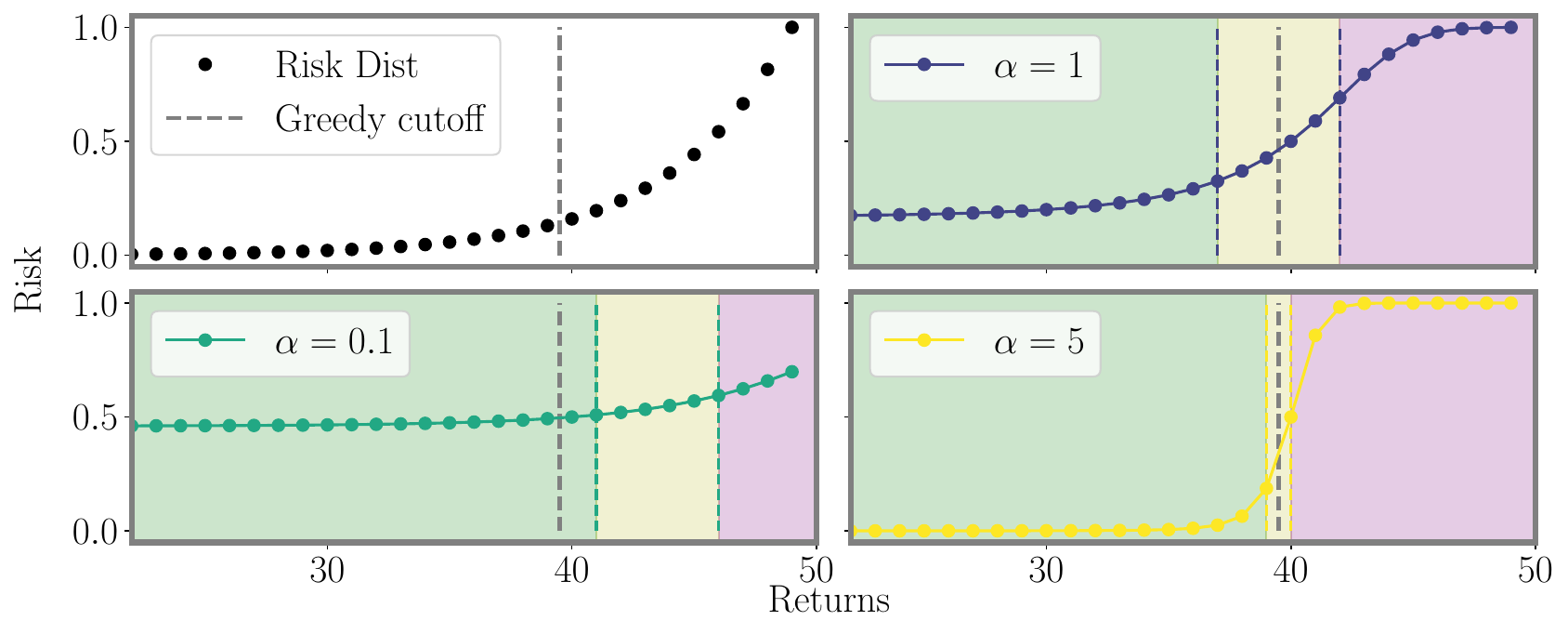}}
    \end{minipage}
    \begin{minipage}{0.1\textwidth}
    \includegraphics[scale=0.08]{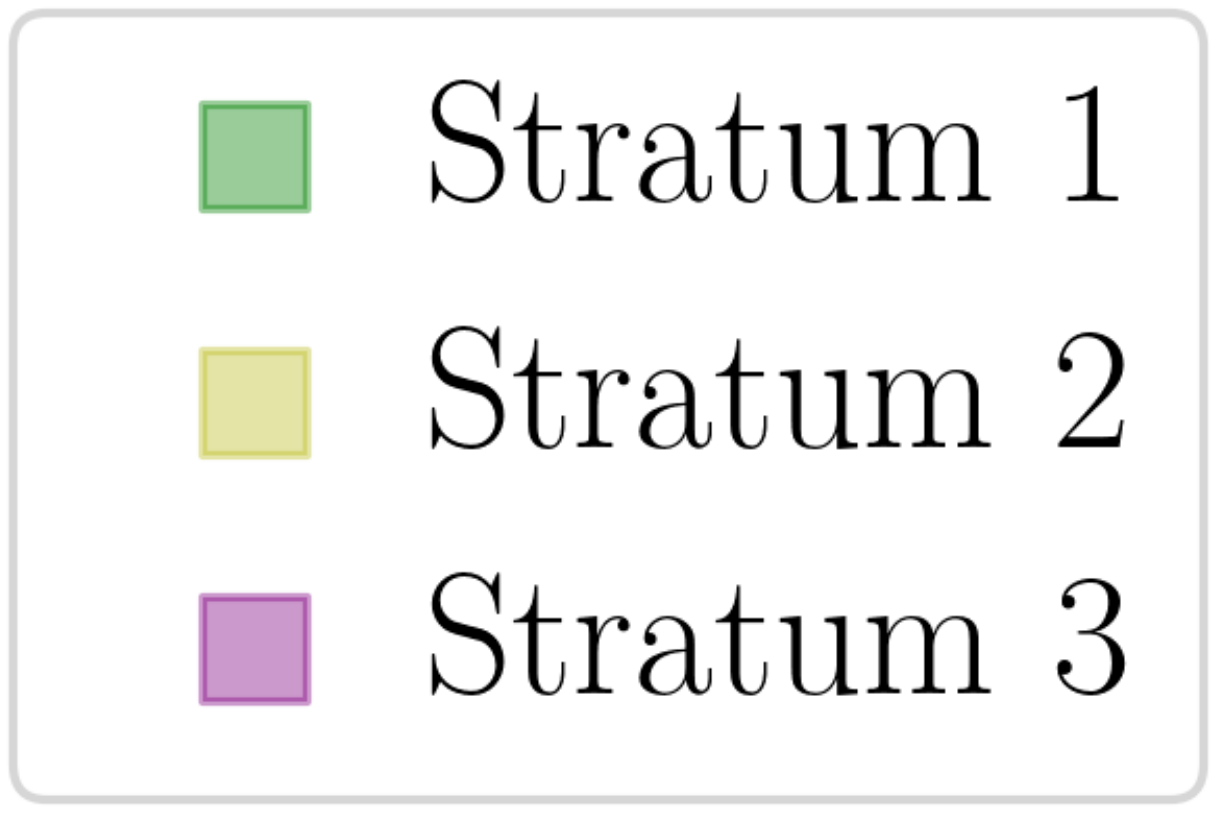}
    \end{minipage}
    \begin{minipage}{0.49\textwidth}
    \subcaptionbox{Probability of sampling an individual}{\includegraphics[width=.75\textwidth]{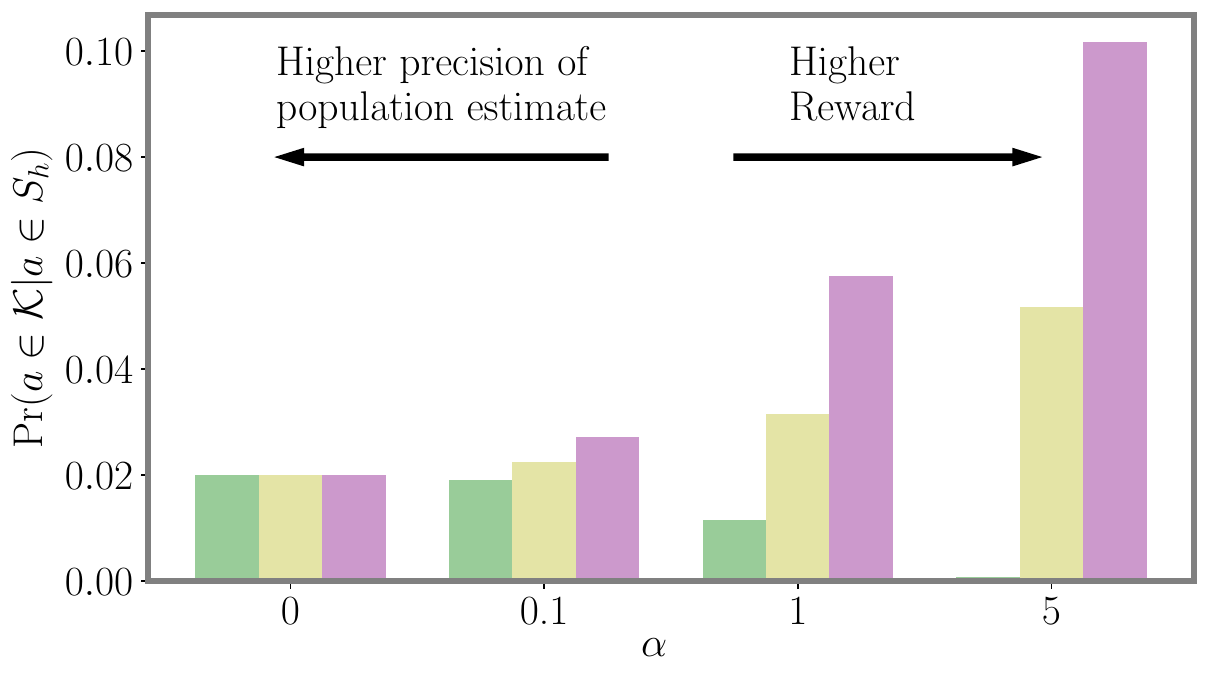}}
    \end{minipage}
    \caption{Illustration of ABS on 50 synthetic observations. (a) Hypothetical risk distribution and three parameterizations corresponding to different values of $\alpha$: 0.1, 1, and 5.    Greedy selection, represented by the dotted (gray) line in each panel would choose the $K=10$ returns with the highest risk. The parameterized risk distributions are clustered into three strata ($S_1$, $S_2$, $S_3$), represented by the colored panels.   As $\alpha$ varies, the cluster assignments change. (b) Probabilities of sampling a single individual from the three strata. As $\alpha$ increases, more weight is put onto the higher risk returns (Stratum 3). 
    }
    \label{fig:abs_sampling}
\end{figure*}

Our NRP stratified random audit sample covers from 2006 to 2014. We use 500 covariates as inputs to the model which are a superset of those currently used for fitting the DIF model.
The covariates we use include every value reported by a taxpayer on a tax return.  For example, the amount reported in Box 9 of Form 1040 is Total income and would be included in these covariates. 
\ifarxiv
Table~\ref{tab:NRP}
\else
Table 5
\fi
, in the Appendix, provides summary statistics of the NRP research audits conducted on a yearly basis. Since NRP audits are stratified, the unweighted means represent the average adjustment made by the IRS to that year's return for all audited taxpayers in the sample. The weighted mean takes into account stratification weights for each sample. One can think of the weighted mean as the average taxpayer misreporting across all taxpayers in the United States, while the unweighted mean is the average taxpayer misreporting in the NRP sample.

\textbf{Problem Formulation.} We formulate the optimize-and-estimate structured bandit problem setting in the scenario where there is an extremely large, but finite, number ($N_t$) of arms ($a \in \arms_t$) to select from at every round. This set of arms is the population at timestep $t$. The population can vary such that the set of available arms may be totally different at every step, similar to a sleeping or volatile bandit~\citep{nika2021contextual}. In fact, it may not be possible to monitor any given arm over several timesteps.\footnote{Note the reason we make this assumption is because the NRP data does not track a cohort of taxpayers, but rather randomly samples. We are not guaranteed to ever see a taxpayer twice.} To make the problem tractable, it is assumed that the reward for a given arm can be modeled by a shared function $r_t^a = f_{\theta^*}(X^a_t)$ where $X_t^a$ are some set of features associated with arm $a$ at timestep $t$, and $\theta^*$ are the parameters of the true reward function. Assume $f \in \mathcal{F}$ is any realizable or $\epsilon$-realizable function. Thus, as is typical of the structured bandit setting ``choosing
one action allows you to gain information about the rewards of other actions''~\citep[p. 301]{lattimore2020bandit}.
The agent chooses a batch of $K_t$ arms to: (1) maximize reward; (2) yield an accurate and unbiased estimate of the average reward across all arms -- even those that have not been chosen (the population reward).
Thus we seek to optimize a selection algorithm that chooses non-overlapping actions $(\hat{a}_1, ..., \hat{a}_K)$ according to a selection policy ($\varpi$) and outputs a population estimate ($\hat{\mu}_\varkappa$) according to an estimation algorithm ($\varkappa$):

\begin{align}
    \min_{\varpi,\varkappa} &~~~\mathbb{E}_{\mathcal{D}} \left[\sum_{t=1}^T \sum_{k=1}^K r^*(a^*_k) - r^*(\hat{a}_k)\right] + \mathbb{V}_{\mathcal{D}, \varpi}\left(\mu^*(t) - \hat{\mu}_\varkappa(t)\right)\\
    \text{s.t.}&~~~ |\hat{\mu}_\varkappa(t) - \mu^*(t)| \rightarrow \mathcal{N}(0, \sigma) \quad \text{as}\quad K \rightarrow N,
\end{align}

where $\mathcal{D}$ is the underlying distribution from which all taxpayers are pulled. In our IRS setting each arm ($a_t$) represents a taxpayer which the policy could select for a given year ($t$). The associated features ($X_t^a$) are the 500 covariates in our data for the tax return. The reward ($r_t^a$) is the adjustment recorded after the audit. 
The population average reward that the agent seeks to accurately model is the average adjustment (summing together would instead provide the tax gap).

\section{Methods}

We focus on three methods: (1) $\epsilon$-greedy; (2) optimism-based approaches; (3) ABS sampling
\ifarxiv
(see Appendix~\ref{app:baselines}
\else
(see Appendix F
\fi
for reasoning and method selection criteria).

\textbf{$\epsilon$-greedy.} Here we choose to sample randomly with probability $\eps$. Otherwise, we select the observation with the highest predicted reward according to a fitted model $f_{\htheta}(X_t^a)$, where $\htheta$ indicates fitted model parameters. To batch sample, we repeat this process $K$ times. The underlying model is then trained on the received observation-reward pairs, and we repeat. For population estimation, we use a model-based approach (see, e.g., \citealt{esteban2019estimating}). After the model receives the true rewards from the sampled arms, the population estimate is predicted as: $\hat{\mu}(t) = \frac{1}{\sum_a w_a}\sum_{a\in \arms_t} w_af_{\htheta}(X_t^a) $, where $w_a$ is the NRP sample weight\footnote{The returns in each NRP strata can be weighted by the NRP sample weights to make the sample representative of the overall population, acting as inverse propensity weights.  We use NRP weights for population estimation. See 
\ifarxiv
Appendix~\ref{app:nrp_weights}.
\else
Appendix K.
\fi
} from the population distribution. 

\textbf{Optimism.} We refer readers to~\citet{lattimore2020bandit} for a general introduction to Upper Confidence Bound (UCB) and optimism-based methods. %
We import an optimism-based approach into this setting as follows. Consider a random forest with $B$ trees $T_1,T_2,\dots,T_B$. We form an optimistic estimate of the reward for each arm according to: $\hat{\rho}_t^a = \frac{1}{B}\sum_b T_b(X_t^a) + Z\Var_b(T_b(X_t^a))$, where $Z$ is an exploration parameter based on the variance of the tree-based predictions, similar to \citet{hutter2011sequential}. We select the $K$ returns with the largest optimistic reward estimates. We shorthand this approach as UCB and use the same model-based population estimation method as $\epsilon$-greedy.

\textbf{ABS Sampling.}   Adaptive Bin Sampling brings together sampling and bandit literatures to guarantee statistically unbiased population estimates, while enabling an explicit trade-off between reward and the variance of the estimate. In essence, ABS performs adjustable \emph{risk-proportional} random sampling over optimized population strata. By maintaining probabilistic sampling, ABS can employ HT estimation to achieve an unbiased measurement of the population.

Pseudocode is given in Algorithm 1. Fix timestep $t$ and let $K$ be our budget. Let $\hat{r}_a=f_{\htheta}(X_t^a)$ be the predicted risk for return $X_t^a$.  First we sample the top $\zeta$ returns. To make the remaining $K-\zeta$ selections,
we parameterize the predictions with a mixing function $\hat{\rho}_a$ intended to smoothly transition focus between the reward and variance objectives, but whose only requirement is that it be monotone (rank-preserving). For our empirical exploration we examine two such mixing functions, a logistic function, $\hat{\rho}_a=\frac{1}{1 + \exp(-\alpha (\hat{r}_a-\kappa))}$ and an exponential function $\hat{\rho}_a = \exp(\alpha \hy_a)$. $\kappa$ is the value of the $K$-th largest value amongst reward predictions $\{\hat{r}_t^a\}$. As $\alpha$ decreases, $\{\hat{\rho}_t^a\}$ approaches a uniform distribution which results in lower variance for $\hmu(t)$ but lower reward. As $\alpha$ increases, the variance of $\hmu(t)$ increases but so too does the reward. Figure~\ref{fig:abs_sampling} provides a visualization of this.

 The distribution of transformed predictions $\{\hat{\rho}_a\}$ is then stratified into $H$ non-intersecting strata $S_1,\dots,S_H$. We choose strata in order to minimize intra-cluster variance, such that there are at least $K-\zeta$ points per bin:
\begin{equation}
\label{eq:min_variance}
    \begin{aligned}
    \min_{S_1,\dots,S_H:\; |S_h|\geq K-\zeta} \quad \sum_{h} \sum_{\hz\in S_h} \norm{\hz - \lambda_h}^2, 
    \end{aligned}
\end{equation}
where $\lambda_h=|S_h|^{-1}\sum_{\hat{\rho}\in S_h}\hz$ is the average value of the points in bin $b$.
We place a distribution $(\pi_h)$ over the bins by averaging the risk in each bin:
\begin{equation}
\label{eq:pi}
    \pi_h = \frac{\lambda_h}{\sum_{h'}\lambda_{h'}}.
\end{equation}
\begin{algorithm}
   \caption{ABS (Logistic)}
   \label{alg:abs}
\begin{algorithmic}
   \STATE {\bfseries Input:} $\alpha$, $H$, $\zeta$, $K$, $(X_0, r_0)$
   \STATE Train model $f_{\htheta}$ on initial data $(X_0,r_0)$. 
   \FOR{$t=1,\dots,T$}
   \STATE Receive observations $X_t$
   \STATE Predict rewards $\hy_a = f_{\htheta}(x_a)$. 
   \STATE Sample top $\zeta$ predictions.
   \STATE $\forall_a$ $\hat{\rho}_a \gets (1 + \exp(-\alpha(\hy_a - \kappa))^{-1}$ 
   \STATE Construct strata $S_1,\dots,S_H$ by solving \eqref{eq:min_variance}. 
   \STATE Form distribution $\{\pi_h\}$ over strata via \eqref{eq:pi}. 
    \REPEAT  
   \STATE $h\sim (\pi_1,\dots,\pi_H)$ 
   \STATE Sample arm uniformly at random from $S_h$.
   \UNTIL{$K-\zeta$ samples drawn}
   \STATE Compute $\hmuht$ once true rewards are collected. 
   \STATE Retrain model $\hat{f}$ on $(\cup_i^t X_i, \cup_i^t r_i)$.  
   \ENDFOR
\end{algorithmic}
\end{algorithm}
To make our selection, we sample $K-\zeta$ times from $(\pi_h,\dots,\pi_H)$ to obtain a bin, and then we sample \emph{uniformly} within that bin to choose the return. We do not recalculate $(\pi_1,\dots,\pi_H)$ after each selection, so while we are sampling without replacement at the level of returns (we cannot audit the same taxpayer twice), we are sampling with replacement at the level of bins. 
The major benefit of ABS is that by sampling according to the distribution $\pi$, we can employ HT estimation to eliminate bias. Indeed, if $\mathcal{K}$ is the set of arms sampled during an epoch,  $\hat{\mu}_{HT}(t) = \frac{1}{\sum_a w_a}\sum_{a\in \mathcal{K}} \frac{w_ar_a}{p_a}$ is an unbiased estimate of the true population, where $p_a$ is the probability that arm $a$ was selected (i.e., $\Pr(a\in\mathcal{K})$) and $w_a$ is the NRP weight. Like with other HT-based methods~\cite{potter1990study,alexander1997making}, to reduce variance we also add an option for a minimum probability of sampling a bin, which we call the trim \%. See 
\ifarxiv
Appendix \ref{app:abs_sampling} 
\else
Appendix N
\fi
for more details, proof of unbiased estimation, and estimator variance. See 
\ifarxiv
Appendix~\ref{app:regret}
\else
Appendix U
\fi
for regret bounds.

\textbf{Reward Structure Models.} As the data is highly non-linear and high-dimensional, we use Random Forest Regression (RFR) for our reward model. We exclude linear models from our suite of algorithms after verifying that they consistently underperform RFR
\ifarxiv
(Appendix~\ref{app:functionapproximator}).
\else
(Appendix M).
\fi
We do not include neural networks in this analysis as the data regime is too small. Future approaches might build on this work using pretraining methods suited for a few-shot context~\citep{bommasani2021opportunities}. We do compare to an LDA baseline
\ifarxiv
(Appendix~\ref{app:lda}).
\else
(Appendix~T.3).
\fi
This is included both as context to our broad modeling decisions, and as an imperfect stylized proxy for one component of the current risk-based selection approach used by the IRS.

\section{Evaluation Protocol}
\label{sec:eval}
We evaluate according to three metrics: cumulative reward, percent difference of the population estimate, and the no-change rate. More details in
\ifarxiv
Appendix~\ref{app:pd} and \ref{app:rare}.
\else
Appendix L.
\fi

\textbf{Cumulative reward ($R$)} is simply the total reward of all arms selected by the agent across the entire time series $\mathbb{E}\left[\left(\sum_{t}^T \sum_k^K r_{a_k}\right)\right]$. It represents the total amount of under-reported tax revenue returned to the government after auditing. This is averaged across seeds and denoted as $R$.

\textbf{Percent difference ($\mu_{PE}$, $\sigma_{PE}$)} is the difference between the estimated population average and the true population average: $100\% * (\hat{\mu} - \mu^*) / \mu^*$. $\mu_{PE}$ is absolute mean percent difference across seeds (bias). $\sigma_{PE}$ is the standard deviation of the percent difference across random seeds.

\textbf{No-change rate ($\mu_{NR}$)} is the percent of arms that yield no reward where we round down such that any reward $<$\$200 is considered no change $\mu_{NR} = \mathbb{E}\left[\left( (1/T) \sum_t^T (1/K) \sum_{k}^K \ind\{r_{a_k} < 200)\}\right]\right)$. NR is of some importance. An audit that results in no adjustment can be perceived as unfair, because the taxpayer did not commit any wrongdoing~\citep{lawsky2008fairly}. It can have adverse effects on future compliance~\citep{nochange,lederman2018does}. $\mu_{NR}$ is the average NR across seeds.

\textbf{Experimental Protocol. } Our evaluation protocol for all experiments follows the same pattern. For a given year we offer 80\% of the NRP sample as arms for the agent to select from.
We repeat this process across 20 distinct random seeds such that there are 20 unique subsampled datasets that are shared across all methods, creating a sub-sampled bootstrap for Confidence Intervals (more in
\ifarxiv
Appendix~\ref{app:confidenceintervals}).
\else
Appendix S).
\fi
Comparing methods seed-to-seed will be the same as comparing two methods on the same dataset.
Each year, the agent has a budget of 600 arms to select from the population of 10k+ arms (description of budget selection in 
\ifarxiv
Appendix~\ref{app:budgets}). 
\else
Appendix~R).
\fi
We delay the delivery of rewards for one year. This is because the majority of audits are completed and returned only after such a delay~\citep{debacker2018effects}. Thus, the algorithm in year 2008 will only make decisions with the information from 2006. Because of this delay the first two years are randomly sampled for the entire budget (i.e., there is a warm start).
After receiving rewards for a given year, the agent must then provide a population estimate of the overall population average for the reward (i.e., the average tax adjustment after audit). This process repeats until 2014, the final year available in our NRP dataset (diagram in
\ifarxiv
Appendix~\ref{app:expsetup2}).
\else
Appendix O).
\fi

\begin{table}
    \centering
    {
    {\bf Best Reward Settings}
    \resizebox{.49\textwidth}{!}{
    \begin{tabular}{ c r l l l l l}
    \toprule 

& \emph{Policy} &  ${R}$ &  $\mu_{PE}$ & $\sigma_{PE}$ & $\mu_{NR}$\\
\midrule 
\multirow{ 4}{*}{Unbiased} & ABS-1 & \textbf{\$41.5M$^*$} & \textbf{0.4} {\color{green}\checkmark} & 31.0  & \textbf{37.6\%} \\
& $\eps$-only & \$41.3M$^*$ &  4.3{\color{green}\checkmark} & 37.4 &  38.3\% \\
& ABS-2 & \$40.5M$^*$ & 0.6{\color{green}\checkmark} & 24.5 &  38.3\%\\
& Random &\$12.7M &  1.5{\color{green}\checkmark} & \textbf{14.7} & 53.1\% \\
\hline
\multirow{ 4}{*}{Biased} &Greedy & \textbf{\$43.6M$^*$} &  16.4  {\color{red}\xmark}& 8.8  & \textbf{36.5\%}  \\
 & UCB-1 & \$42.4M$^*$ & 15.3 {\color{red}\xmark} & 9.4 & 38.6\%  \\
& $\epsilon$-Greedy& \$41.3M$^*$ &  \textbf{6.1} {\color{red}\xmark} & \textbf{7.5} &  38.3\% \\
 & UCB-2 & \$40.7M$^*$& 15.6 {\color{red}\xmark} & 10.21 & 40.7\% \\
\end{tabular}}}

    \caption{
    Best settings with overlapping CIs ($^*$) on $R$.
    $R$ is a cumulative reward. $\mu_{PE}$ is the average percent difference of the population estimate across seeds. $\sigma_{PE}$ is the standard deviation of the percent difference across seeds. $\mu_{NR}$ is the no change rate. 
    Extended table with hyperparameters for all displayed methods is in
    \ifarxiv
    Appendix~\ref{app:allresults}, 
    \else
    Appendix~T, 
    \fi
    selection method in
        \ifarxiv
    Appendix~\ref{app:hyperpararmeters}.
    \else
    Appendix~P.
    \fi
 Biased methods with no guarantees are highly undesirable ({\color{red}\xmark}).
    $\eps$-only is the same as $\eps$-Greedy, but population estimation uses only the $\eps$ sample as a random sample. Random is where the full, 600 arm, sample is random.
    }
    \label{tab:best_reward}
\end{table}

\section{Results}

\begin{figure*}[!htbp]
    \centering

    \includegraphics[width=.48\textwidth]{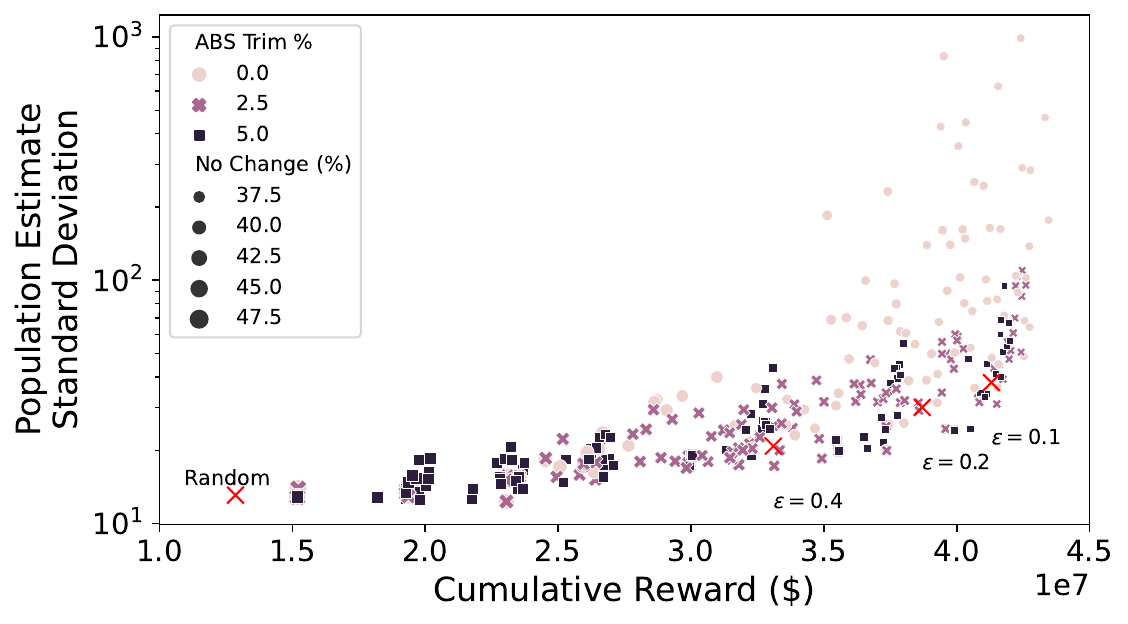}
    \hfill
    \includegraphics[width=.48\textwidth]{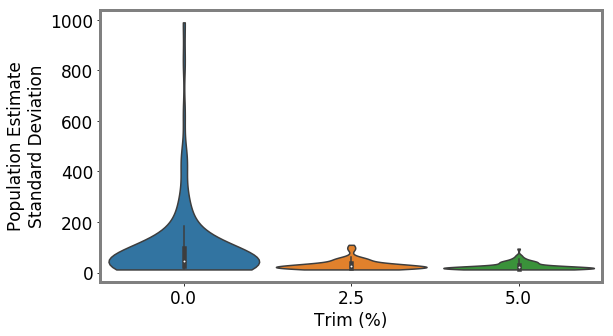}
    \includegraphics[width=.32\textwidth]{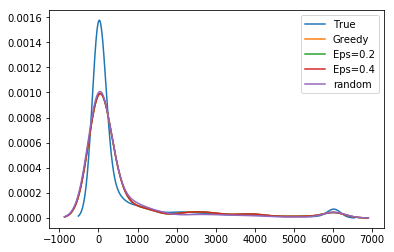}
    \includegraphics[width=.32\textwidth]{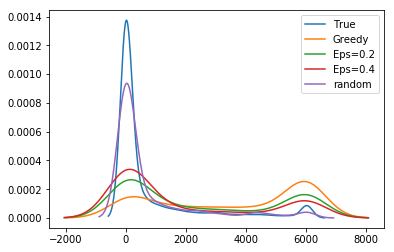}
        \includegraphics[width=.32\textwidth]{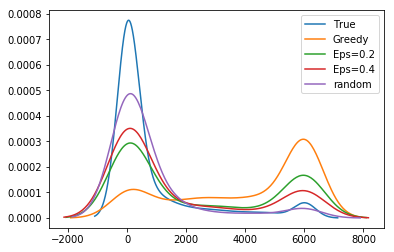}
    \caption{(Top Left) Population estimation empirical standard deviation versus reward for a grid of ABS hyperparameters.  ({\color{red}x}) and associated labels indicate $\epsilon$-only and fully random sample. (Top Right) Population estimation variance as a function of the trim. (Bottom) A kernel density plot of the distribution of sampled arms from 2006 (top left) to 2014 (bottom). X-axis is true reward. Y-axis is sampling distribution density.}
    \label{fig:variance}

\end{figure*}
We highlight several key findings with additional results and sensitivity analyses in
\ifarxiv
Appendix~\ref{app:allresults}.
\else
Appendix~T.
\fi

\textbf{Unbiased population estimates are possible with little impact to reward.} ABS sampling can achieve similar returns to the best performing methods in terms of audit selection, while yielding an unbiased population estimate (see Table~\ref{tab:best_reward}). 
Conversely, greedy, $\epsilon$-greedy, and UCB approaches -- which use a model-based population estimation method -- do not achieve unbiased population estimates. 
Others have noted that adaptively collected data can lead to biased models~\citep{nie2018adaptively,neel2018mitigating}. 
In many public sector settings provably unbiased methods like ABS are \emph{required}.
For $\eps$-greedy, using the $\eps$-sample only would also achieve an unbiased estimate, yet due to its small sample size the variance is prohibitively high.
ABS reduces variance by 16\% over the best $\eps$-only method, yielding even better reward. Trading off \$1M over 9 years improves variance over $\eps$-Greedy ($\eps$-only) by 35\%. It is possible to reduce this variance even further at the cost of some more reward (see Figure~\ref{fig:variance}). Note, due to an extremely small sample size, though the $\eps$ sample is unbiased in theory, we see some minor bias in practice.
Model-based estimates are significantly lower variance, but biased. This may be because models re-use information across years, whereas ABS does not. Future research could re-use information in ABS to reduce variance, perhaps with a model's assistance.
Nonetheless, we emphasize that model-based estimates without unbiasedness guarantees are unacceptable for many public sector uses from a policy perspective.

\textbf{ABS allows fine-grained control over variance-reward trade-off.} We sample a grid of hyperparameters for ABS (see 
        \ifarxiv
    Appendix~\ref{app:hyperpararmeters}).
    \else
    Appendix~P).
    \fi
Figure~\ref{fig:variance} shows that more hyperparameter settings close to optimal rewards have higher variance in population estimates. We can control this variance with the trimming mechanism. This ensures that each bin of the risk distribution will be sampled some minimum amount. Figure~\ref{fig:variance} also shows that when we add trimming, we can retain large rewards and unbiased population estimates. Top configurations (Table~\ref{tab:best_reward}) can keep variance down to only 1.7x that of a random sample, while yielding 3.2x reward. 
While $\epsilon$-greedy with the random sample only does surprisingly well, optimal ABS configurations have a better Pareto front. We can fit a function to this Pareto front and estimate the marginal value of the reward-variance trade-off (see
\ifarxiv
Appendix~\ref{app:pricemetric}).
\else
Appendix~T.2).
\fi

\begin{figure*}[!htbp]
    \centering

        \includegraphics[width=\textwidth]{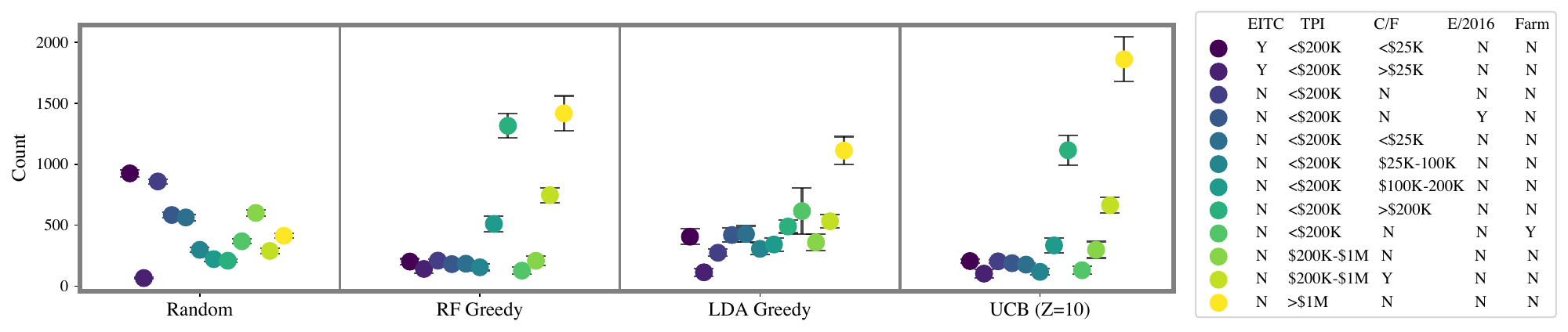}
  \includegraphics[width=.3\textwidth]{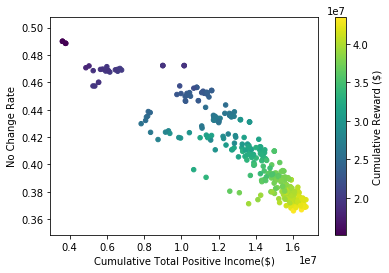}
    \includegraphics[width=.4\textwidth]{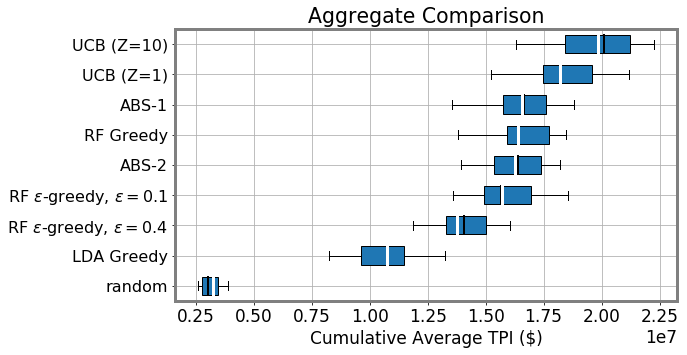}
    \caption{(Top) The distribution of audit classes for several approaches. Markers follow
the same order as the legend from left to right. EITC Stands for ``Earned Income Tax Credit'' and ``TPI'' for Total Positive Income. (Bottom Left) Distribution of TPI for ABS hyperparameter settings. (Bottom Right) Distribution of TPI for various methods. }
    \label{fig:income}
\end{figure*}

\textbf{Greedy is not all you need.} Greedy surprisingly achieves more optimal reward compared to all other methods (see Table~\ref{tab:best_reward}). This aligns with prior work suggesting that a purely greedy approach in contextual bandits might be enough to induce sufficient exploration under highly varied contexts~\citep{bietti2018contextual,kannan2018smoothed,bastani2021mostly}. Here, there are several intrinsic sources of exploration that may cause this result: intrinsic model error, covariate drift
\ifarxiv
(see Appendix Table~\ref{tab:NRP}), 
\else
(see Appendix Table~5), 
\fi
differences in tax filing compositions, and the fact that our population of arms already come from a stratified random sample (changing in composition year-to-year).

Figure~\ref{fig:variance} (bottom) demonstrates greedy sampling's implicit exploration for one random seed. As the years progress, greedy is (correctly) more biased toward sampling arms with high rewards. Nonetheless, it yields a large number of arms that are the same as a random sample would yield. This inherent exploration backs the hypothesis that the test sample is highly stochastic, leading to implicit exploration. It is worth emphasizing that in a larger population and with a larger budget, greedy's exploration may not be sufficient and more explicit exploration may be needed.
The key difference from our result and prior work showing greedy's surprising performance~\cite{bietti2018contextual,kannan2018smoothed,bastani2021mostly} is our additional population estimation objective. The greedy policy has a significant bias when it comes to model-based population estimation. 
This bias is similar -- but not identical -- to the bias reported in other adaptive data settings \citep{thrun1993issues,nie2018adaptively,shin2021bias,farquhar2021statistical}.
Even a 10\% random sample -- significantly underpowered for typical sampling-based estimation -- can reduce this bias by more than 2.5$\times$ (see Table~\ref{tab:best_reward}). Even if greedy can be optimal for a high-variance contextual bandit, it is not optimal for the optimize-and-estimate setting. $\epsilon$-greedy achieves a compromise between variance that may be more acceptable in settings when some bias is permitted, but bias is not desirable in most public sector settings.
We also show that RFR regressors significantly outperform LDA and that incorporating non-random data helps 
\ifarxiv
(Appendix~\ref{app:lda}). 
\else
(Appendix T.3). 
\fi
This is a stylized proxy of the status quo system that uses a small $\eps$-only sample (NRP) for population estimates and an LDA-like algorithm (DIF) for selection.

\textbf{A more focused approach audits higher cumulative total positive income.} A key motivator for our work is that inefficiently-allocated randomness in audit selection will not only be suboptimal for the government, but could impose unnecessary burdens on taxpayers~\citep{lawsky2008fairly,davis2012unequal}. An issue that has received increasing attention by policymakers and commentators in recent years concerns the fair allocation of audits by income~\citep{kiel,irs2021update,treasury2021american}. Although we do not take a normative position on the precise contours of a fair distribution of audits, we examine how alternative models shape the income distribution of audited taxpayers.

As shown in Figure~\ref{fig:income}, we find that as methods become more optimal we see an increase in the total positive income (TPI) of the individuals selected for audit (RF Greedy selects between \$1.8M and \$9.4M more cumulative TPI than LDA Greedy, effect size 95\% CI matched by seed). We also show the distribution of ABS hyperparameter settings we sampled. As the settings are more likely to increase reward and decrease no change rates, the cumulative TPI increases. This indicates that taxpayers with lower TPI are less likely to be audited as models are more likely to sample in the higher range of the risk distribution. We confirm this in  Figure~\ref{fig:income} (top) which shows the distribution of activity classes sampled by different approaches. These classes are used as strata in the NRP sample. The UCB and RF Greedy approaches are more likely to audit taxpayers with more than \$1M in TPI (with UCB sampling this class significantly more, likely due to heteroskedasticity). More optimal approaches also significantly sample those with $<$\$200K in TPI, but more than \$200K reported on their Schedule C or F tax return forms (used to report business and farm income, respectively).

\textbf{Errors are heteroskedastic, causing difficulties in using model-based optimism methods.} Surprisingly, our optimism-based approach audits tax returns with higher TPI more often (\$1.2M to \$5.8M million cumulative TPI more than RF Greedy) despite yielding similar returns as the greedy approach. We believe this is because adjustments and model errors are heteroskedastic.
Though TPI is correlated with the adjustment amount (Pearson $r=0.49, p<10^{-5}$), all errors across model fits were heteroskedastic according to a Breusch–Pagan test ($p < 10^{-5}$). 
A potential source of large uncertainty estimates in the high income range could be because: (1) there are fewer datapoints in that part of the feature space; (2) NRP audits may not give an accurate picture of misreporting at the higher part of the income space, resulting in larger variance and uncertainty~\citep{guyton2021tax}; or (3) additional features are needed to improve precision in part of the state space. This makes it difficult to use some optimism-based approaches since there is a confound between aleatoric and epistemic uncertainty.
As a result, optimism-based approaches audit higher income individuals more often, but do not necessarily achieve higher returns. 
This poses another interesting challenge for future research.

\section{Discussion}

We have introduced the optimize-and-estimate structured bandit setting. The setting is motivated by common features of public sector applications (e.g., multiple objectives, batched selection), where there is wide applicability of sequential decision making, but, to date, limited understanding of the unique methodological challenges. 
We empirically investigate the use of structured bandits in the IRS setting and show that ABS conforms to IRS specifications (unbiased estimation) and enables parties to explicitly trade off population estimation variance and reward maximization.
This framework could help address longstanding concerns in the real-world setting of IRS detection of tax evasion. It could shift audits toward tax returns with larger understatements (correlating with more total positive income) and recover more revenue than the status quo, while maintaining an unbiased population estimate. Though there are other real-world objectives to consider, such as the effect of audit policies on tax evasion, our results suggest that unifying audit selection with estimation may help ensure that processes are as fair, optimal, and robust as possible. 
We hope that the methods we describe here are a starting point for both additional research into sequential decision-making in public policy and new research into optimize-and-estimate structured bandits.

\section*{Acknowledgements}
We would like to thank Emily Black, Jason DeBacker, Hadi Elzayn, Tom Hertz, Andrew Johns, Dan Jurafsky, Mansheej Paul, Ahmad Qadri, Evelyn Smith, and Ben Swartz for helpful discussions. This work was supported by the Hoffman Yee program at Stanford’s Institute for Human-Centered Artificial Intelligence and Arnold Ventures. PH is supported by the Open Philanthropy AI Fellowship. This work was conducted while BA was at Stanford University. The findings, interpretations, and conclusions expressed in this paper are entirely those of the authors and do not necessarily reflect the views or the official positions of the U.S. Department of the Treasury or the Internal Revenue Service. Any taxpayer data used in this research was kept in a secured Treasury or IRS data repository, and all results have been reviewed to ensure no confidential information is disclosed.

\bibliography{main}

\ifarxiv

\onecolumn

\appendix

\section{Software and Data}
\label{app:code}

We are unable to publish even anonymized data due to statutory constraints. 26 U.S. Code \S~6103. All code, however, is available at \url{https://github.com/reglab/irs-optimize-and-estimate}. We also provide datasets that can act as rough proxies to the IRS data for running the code, including the 
Public Use Microdata Sample (PUMS)~\citep{pums} and Annual Social and Economic Supplement (ASEC) of the Current Population Survey (CPS)~\citep{asec}. These two datasets are provided by the U.S. Census Bureau. In this case, we use the proxy goal of identifying high-income earners with non-income-based features while maintaining an estimate of total population average income.

\section{Carbon Impact Statement}
\label{app:carbon}

As suggested by \citet{lacoste2019quantifying}, \citet{henderson2020towards}, and others, we report the energy and carbon impacts of our experiments. While we are unable to calculate precise carbon emissions from hardware counters, we give an estimate of our carbon emissions. We estimate roughly 12 weeks of CPU usage total, including hyperparameter optimization and iteration on experiments, on two Intel Xeon Platinum CPUs with a TDP of 165 W each. This is equal to roughly 665 kWh of energy used and 471 kg CO$_{2eq}$ at the U.S. National Average carbon intensity.

\section{Importance and Relevance of the Optimize-and-Estimate Setting}
\label{app:relevance}

We note that the optimize-and-estimate setting is essential to many real-world tasks. First, we highlight that many Federal Agencies are bound by law to estimate improper payments under the Improper Payments Information Act of 2002 (IPIA), as amended by the Improper Payments Elimination and Recovery Act of 2010 (IPERA) and the Improper Payments Elimination and Recovery Improvement Act of 2012 (IPERIA). ``An improper payment is any payment that should not
have been made or that was made in an incorrect amount
under statutory, contractual, administrative, or other
legally applicable requirements.''\footnote{\url{https://www.ignet.gov/sites/default/files/files/19\%20FSA\%20IPERA\%20Compliance\%20Slides.pdf}}
OMB guidance varies year-to-year on how improper payments should be estimated, but generally they must be ``statistically valid'' (unbiased estimate of the mean)~\citep{omb2018requirements}. In past years OMB has also required tight confidence intervals on estimates~\citep{omb2018requirements}.

Generally this means that agencies will need to conduct audits, as the IRS does described in this paper, to determine if there was misreporting or improper payments were made. Effectively, the optimize-and-estimate problem that we highlight here can apply more broadly to any federal agency that falls under the laws listed above.

We also highlight that the optimize-and-estimate structured bandit setting is not uniquely important to the public sector. Private sector settings also have all the qualities of an optimize-and-estimate problem. We consider one such application below: content moderation.

During some time period, a platform will have a large set of content that might violate their policies.
They will have a set of moderators who will audit content that should be taken down.
There will likely be more content in need of moderation than there are moderators. As a result, it is important to take down the most egregious cases of policy-violating content, while assessing an overall estimate of prevalence on the platform.
To optimize this process the platform could construct an optimize-and-estimate problem as we do here. In this case, each arm would be a piece of content that needs review. Reward is a rating on how offensive or egregious the content policy violation was. The population estimate would then give an estimate of prevalence and valence of content policy violations on the platform. Note, here unbiasedness is likely equally important to the platform since a heavy bias will incorrectly affect policy decisions about content moderation.

\section{Society/Ethics Statement}
\label{app:ethics}

As part of the initial planning of this collaboration, the project proposal was presented to an Ethics Review Board and was reviewed by the IRS. While the risks of this retrospective study -- which uses historical data -- are minimal, we are cognizant of distributive effects that targeted auditing may have. 
In this work we examine the distribution of audits across income, noting that more optimal models audit higher income taxpayers -- in line with current policy proposals for fair tax auditing~\citep{treasury2021american}. Our collaboration will also be investigating other notions of fairness in separate follow-on and concurrent work as they require a more in-depth examination than can be done in this work alone.

There are multiple important (and potentially conflicting) goals in selecting who to audit, including maximizing the detection of under-reported tax liability, maintaining a population estimate, avoiding the administrative and compliance costs associated with false positive audits, and ensuring a fair distribution of audits across taxpayers of different incomes and other groups. It is important to note that the IRS and Treasury Department will ultimately be responsible for the policy decision about how to balance these various objectives. We see an important contribution of our project as understanding these trade-offs and making them explicit to the relevant policy-makers. We demonstrate how to quantify and incorporate these considerations into a multi-objective model. We also formalize an existing de facto sequential decision-making (SDM) problem to help identify relevant fairness frameworks and trade-offs for policymakers~\citep{henderson2021}. We note, however, that we do not consider a number of other objectives important to the IRS, including deterring tax evasion. 

Finally, as is also well-known, there is no single solution for remedying fairness -- different fairness definitions are contested and mutually incompatible. For this reason, our plan is not to adopt a single, fixed performance measure. Rather, we seek to show how the optimal algorithm varies based on the relative importance one attaches to the alternative goals. For example, in this work we examine how reward-optimal models shift auditing resources toward higher incomes or particular NRP audit classes.
We also note that there are some challenges in examining sub-group fairness, however, including that the IRS may not collect or possess information about protected group status (e.g., race / ethnicity). If such data were available, now-standard bias assessments and mitigation could be implemented. We note that the legality of bias mitigation remains uncertain~\citep{xiang2020reconciling,ho2020affirmative}. Overcoming these challenges requires its own examination.

\ifnta
We note that all models used by the IRS go through extensive internal review processes with robustness and generalizability checks beyond the scope of our work here. No model in this work will be directly used in any auditing mechanism.
\else
With thorough additional evaluation and safeguards, as part of our ongoing collaboration with the IRS, the results of this and related research are currently being incorporated into the continual improvement of the IRS audit selection method.
However, 
we note that all models used by the IRS go through extensive internal review processes with robustness and generalizability checks beyond the scope of our work here. No model in this work will be directly used in any auditing mechanism.
\fi
There are strict statutory rules that limit the use and disclosure of taxpayer data. All work in this manuscript was completed under IRS credentials and federal government security and privacy standards. All authors that accessed data have undergone a background check and been onboarded into the IRS personnel system under the Intergovernmental Personnel Act or the student analogue. That means all data-accessing authors took all trainings on security and privacy related to IRS data and were bound by law to relevant privacy standards (e.g., The Privacy Act of 1974, The Taxpayer Browsing Protection Act, and IRS Policy on Accessing Tax Information).
Consent for use of taxpayer data by IRS for tax administration purposes is statutorily provided for under 26 U.S.C.~\S 6103(h)(1), which grants authority to IRS employees to access data for tax administration purposes.
This manuscript and associated data was cleared under privacy review. All work using taxpayer data was done on a secure system with separate hardware. No taxpayer names were associated with the features used in this work.

\section{Related Work}
\label{sec:related_work}

Please see Table~\ref{tab:rwork} for a brief summary on what sets apart our setting from others and a description of related work below.

\begin{table*}
    \centering
    \resizebox{\textwidth}{!}{
    \begin{tabular}{|c|c|c|c|c|c|c|c}
    \toprule
    Setting & Papers & Batched & Estimation & Volatile Arms  & Per-arm Context & Non-linear  \\
    \midrule
          Multi-armed Bandit   %
          & \citet{deliu2021efficient} & N & Y  & N & N & N \\
          & \citet{caria2020adaptive} & N & Y  & N & N & N \\
          & \citet{kasy2021adaptive} & N & Y  & N & N & N \\
          & \citet{pandey2006handling} & Y & Y  & N & N & N \\
          & \citet{xu2013estimation} & N & Y & N & N & N \\
          & \citet{guo2021learning} & N & Y & N & N & N \\
          \hline
         Contextual Bandit & \citet{dimakopoulou2017estimation} & N & Y  & N & N & N\\
          & \citet{qin2022adaptivity} & N & Y  & N & N & N\\
         & \citet{sen2021top} & Y & N & N & N & Y \\
         & \citet{simchi2020bypassing} & N & N & N & N & Y\\
         & \citet{huang2016linear} & Y & N & N & N & N\\
         \hline
         Structured Bandit &\citet{abbasi2011improved} & N & N & Y & Y & N \\
         & \citet{joseph2018meritocratic} & Y & N & Y & Y & N \\
         \hline
         Ours & - & Y & Y  &Y&Y & Y  \\
         \bottomrule
    \end{tabular}}
    \caption{For clarity, we provide a comparison against related work.}
    \label{tab:rwork}
\end{table*}

\subsection{Similar Applications}

There is growing interest in the application of ML to detecting fraudulent financial statements \citep{dickey2019machine,bertomeu2021using}. Previous methods have included unsupervised outlier detection \citep{de2018tax}, decision trees \citep{kotsiantis2006predicting}, and analyzing statements with NLP \citep{sifa2019towards}. Closer to our methodology is a bandit approach is used by \citet{soemers2018adapting} to detect fraudulent credit card transactions.
Meanwhile, \citet{zheng2020ai} propose reinforcement learning to learn optimal tax policies, but do not focus on enforcement. 
Finally, some work has investigated the use of machine learning for improved audit selection in various settings \citep{howard2020can,ash2021machine,mittal2018bogus}. None of these approaches takes into account population estimation and some do not use sequential decision-making.

\subsection{Multi-objective Decision-making}
Some prior work has investigated general multi-objective optimization in the context of bandits~\cite{drugan2014scalarization,tekin2018multi,turgay2018multi}. Most work in this vein generalizes reward scalars to vectors, and seeks pareto optimal solutions. These techniques do not extend readily to our setting, which has a secondary objective of a particular form (unbiased estimation).

\subsection{Batch Selection}
Other works have examined similar batched selection mechanisms such as the linear structured bandit~\citep{mersereau2009structured,abbasi2011improved,joseph2018meritocratic}, top-k extreme contextual bandit~\citep{sen2021top}, or contextual bandit with piled rewards~\citep{huang2016linear}.

An alternative view of this problem is as a contextual bandit problem with no shared context, but rather a per arm context. This is similar to the setup to the contextual bandit formulation of \citet{li2010contextual} used for news recommendation systems. 
However, unlike in \citet{li2010contextual}, rewards here would have to be delivered after $K$ rounds of selection (where $K$ is the budget of audits that can be selected in a given year). Since the IRS does not conduct audits on a rolling basis, the rewards are delayed and updated all at once. This is similar to the ``piled-reward'' variant of the contextual bandit framework discussed by \citet{huang2016linear} or possibly a variant of contextual bandits with knapsacks~\citep{agrawal2015linear}. 

Notably a large difference in our setting is the scale of the problem (there are 200M+ arms per timestep in the fully scaled problem), the non-linearity of the structure. In other batched settings, it is typically to select K actions given one context, not a per-arm context as is the case for our setting.

\subsection{Inference}

Due to the well-known bias exhibited by data collected by bandit algorithms~\citep{shin2021bias,nie2018adaptively,xu2013estimation}, a large body of work seeks to improve hypothesis testing efficiency and accuracy via an active learning or structured bandit process~\citep{kato2020efficient,mukherjee2020generalized,deliu2021efficient}.
There is also a body of work that seeks to improve estimation and inference properties in bandit settings \citep{kasy2021adaptive,wortman2007maintaining,zhang2020inference,lansdell2019rarely}. For instance, \citet{dimakopoulou2017estimation} consider balancing techniques to improve inference in non-parametric contextual bandits. \citet{chugg2021reconciling} seek to give unbiased population estimates after data has been sampled with a MAB algorithm. \citet{guo2021learning} study how to develop estimation strategies when given the learning process of another low-regret algorithm. 
Some of this work can be classified as click-through-rate estimation~\citep{pandey2006handling,wortman2007maintaining,xu2013estimation}.
Like other adaptive experimentation literature, these works are in the multi-armed bandit (MAB) or contextual bandit settings which do not neatly map onto our own setting. Our work deals with the unique challenges of the IRS setting, requiring the use of optimize-and-estimate \emph{structured} bandits, discussed below.

Some work, similarly to our approach, explicitly considers trading off maximizing reward with another objective~\cite{liu2014trading,rafferty2018bandit}. \citet{caria2020adaptive} develop a Thompson sampling algorithm which trades off estimation of treatment effects with estimation accuracy. \citet{erraqabi2017trading} develop an objective function to trade off rewards and model error. \citet{deliu2021efficient} develop a similar approach for navigating such trade-offs. But all of these works occur in the MAB setting, however, and are difficult to apply to our structured bandit setting. That is because we are not selecting between several treatments, but rather we are selecting a batch of arms to pull which correspond to their own context. Additionally arms are volatile, we do not necessarily know which arm corresponds to an arm in a previous timestep. Finally, arms must be selected in batches with potentially delayed reward. For example, take the Thompson sampling approach of \citet{caria2020adaptive}. In that setting the authors selected among three treatments and examined the treatment effects among them given some context. Yet, in our setting we can never observe any effect for unselected arms so our setting must instead be formulated as a structured bandit.

\subsection{Active Learning}

We also note that there are some similarities between our formulation and the active learning paradigm~~\cite{settles2009active}. For instance, the tax gap estimation requirement could be formulated as a pool-based active learning problem, wherein the model chooses each subsequent point in order to improve its estimation of the tax gap.
This also coincides somewhat with the bandit exploration component since a better model will allow the agent to select the optimal arm more frequently. 
However, the revenue maximization objective we introduce corresponds with the exploit component of the bandit problem and is not found in the active learning framework.

\subsection{Discussion}

Overall, to our knowledge the optimize-and-estimate structured bandit setting has not been proposed as we describe it here. And, more importantly, no one work has examined the unique challenges of the audit selection application in the IRS in as an optimize-and-estimate structured bandit. 
The assumptions we make (essential to most policy contexts including the IRS) differ from each of these other contexts: (1) arms are volatile and cannot necessarily be linked across timesteps;\footnote{Note the reason we make this assumption is because the NRP data does not track a cohort of taxpayers, but rather randomly samples. We are not guaranteed to ever see a taxpayer twice.} (2) decisions are batched; (3) contexts are per arm; (4) the underlying reward structure is highly non-linear; (5) an unbiased population estimate must be retained. These key features also distinguish the optimize-and-estimate structured bandit from past work handling dual optimization and estimation objectives. 
To simplify things, one might think of our work as a top-$k$ contextual bandit, but the key difference is that arms are volatile in our case (you may never see the same arm twice). Thus, we must formulate our setting in a different way: a structured bandit. Thus, perhaps closest to our own is the work of \citet{joseph2018meritocratic} and \citet{abbasi2011improved}. However, we require non-linearity and batched selection, as well as adding the novel estimation objective to this selection setting. These features differentiate it from prior MAB or even contextual bandit work. To our knowledge, this is unlike any of the prior work that handles estimation trade-offs and is the reason why we call the novel domain an optimize-and-estimate structured bandit.
Perhaps closest to our own is the work of \citet{joseph2018meritocratic} and \citet{abbasi2011improved}. But we extend this work to the batched non-linear setting with a unique estimation objective.

\section{Baseline Methods Selection}
\label{app:baselines}

In determining which baselines to use we surveyed existing literature on what methods might be directly applicable. First, we noticed that existing literature that handles estimation problems, such as the adaptive RCT literature~\citep{caria2020adaptive}, do not neatly map onto our setting. We do not have multiple treatments and we never see reward for unselected arms. This ruled out simple methods related to multi-armed bandits. Instead, the closest literature is the structured bandit or linear bandit literature where each arm is assumed to have a context and the policies selects arms as such. Few sampling methods guarantee unbiased estimation in the structured bandit setting, so we turn to a simple $\epsilon$-greedy baseline for unbiasedness. In many ways this is similar to the current IRS setting. The $\epsilon$ sample can be thought of as the NRP sample and then the greedy sample can be thought of as Op audits. This is a natural baseline to compare against. Then, we selected one optimism based approach (UCB) which has proven regret bounds in the linear bandit setting~\citep{lattimore2020bandit}. We show in another setting that a formulation of Thompson sampling does not appear to work well in this structured bandit setting, as seen in Appendix~\ref{app:otherexps}. We do not select more such approaches because model-based approaches are not guaranteed to be unbiased. Thus, while we include one such approach for comparison and analysis, we instead focused our efforts elsewhere instead of adding additional methods that do not meet our optimization criteria. Note, in many ways ABS sampling bears a resemblance to Thompson sampling and we encourage future work to explore more direct mappings of existing literature to the optimize-and-estimate structured bandit setting. For convenience, we make a comparison of this in Table~\ref{tab:rwork}.

\section{Evaluation on Other Datasets}
\label{app:otherexps}

We provide a small ablation study on the additional Current Population Survey (CPS) dataset. We perform the same preprocessing as \citet{chugg2022entropy} and have a goal of predicting the income of a person based on 122 other features. We reuse the optimal ABS hyperparameters from the IRS setting and find similar results on this new dataset. Since CPS is more stationary than the IRS data, the reward-optimal method changes from greedy to UCB. This flows naturally from \citet{bastani2021mostly}, since IRS data is more stochastic, greedy is more optimal. CPS is less stochastic, so UCB is more optimal. ABS remains unbiased and yields high reward. We also implement a version of Thompson sampling where we estimate the standard deviation and mean of the random forest as with the UCB setting, but then sample randomly from a Gaussian distribution with these parameters. We run all permutations for 20 random seeds. We also find that if we try to back out propensity weights for the Thompson sampling approach via Monte Carlo simulations, they are not well-formed. We rolled out 1000 times per timestep and found that even under this regime roughly 75.3\% of the propensity scores were 0. As such, Thompson sampling cannot maintain unbiasedness on its own with an HT-like estimator.

\begin{table}[]
    \centering
    \begin{tabular}{c|c|c|c}
    \toprule
Method & Reward & Bias & Variance\\
\midrule
UCB-1 & 473 & 13.4 & 14.4\\
ABS-2 & 444 & -1.3 & 26.3\\
ABS-1 & 431 & 0.2 & 26.1\\
Thompson & 427.2 & 9.7 & 12.8\\
E-greedy, e=0.4 (Model-based) & 313.5 & 7.25 & 10.6\\
E-greedy, e=0.2 (Model-based) & 243.9 & 7.46 & 10.0\\
\bottomrule
    \end{tabular}
    \caption{Results on the CPS dataset, re-using same optimal hyperparameters as Table 1 in the main text.}
    \label{tab:other_results}
\end{table}

\section{Limitations}
\label{app:limitations}

While we have already expressed several limitations throughout this work, we gather them here as they are fertile ground for research in the \emph{optimize-and-estimate} setting.
First, the ABS approach does not re-use information year-to-year for population estimation. There may be better ways to re-use information in a mixed model-based and model-free mechanism, that retains unbiasedness.
Second, we focus heavily on empirical analysis as we believed it was essential for a setting as important as the IRS. However, future work can delve into theoretical aspects of the problem, potentially examining whether there are lower-variance model-based approaches that retain unbiasedness guarantees.
Third, we note that our evaluations may scale differently to larger datasets. For example, with more data, utilizing the entirety of all audit outcomes, more complicated models might be more feasible.

As an initial work introducing a highly relevant setting and application, we focused more on analysis of different approaches. We have one initial approach, ABS, that meets policy requirements and outperforms baselines. Future work may seek to improve on the performance of this method, focusing more on one novel method rather than analysis, which is our goal.

\subsection{Deterrence}
\label{app:deterrence}
We make several observations about deterrence. First, we do not formally take into account deterrence in this setting, as that would require economic modeling that we believe is outside the scope of this work. We note, however, that economic models of deterrence generally impose relatively strong assumptions about taxpayer knowledge of the audit rate and conventional deterrence models trade off the probability of detection and sanction as policy levers. See, e.g., \citet{slemrod2019tax,snow2005tax}. There is relatively limited evidence about the taxpayer knowledge of audit probabilities and existing empirical literature remains mixed about the deterrence impact of audits~\citep{dularif2019deterrence}. This leads to questions like whether the optimal approach is to use tax rates, and not audits, as policy levers for deterrence, and  this is what makes direct incorporation of deterrence particularly challenging but subject to much fruitful future work. Second, broad based deterrence requires some stochasticity to create uncertainty, as otherwise one part of the population might know they will not be audited and stop complying. Indeed, this is in part why the IRS has historically withheld details of previous algorithmic approaches. We note that the ABS trim parameter ensures coverage of the population and that every part of the population has some chance of being audited (though it is minimized to the extent possible for low-risk parts of the distribution). In that sense, the additional weight on exploration induced by ABS / population estimation could promote general deterrence. In short, the explore-exploit tradeoff is quite relevant for deterrence, but the economic modeling required is beyond the scope of this work.

\section{Covariate Drift}
\label{app:drift}

We also characterize the covariate drift year to year. We calculate the average per-covariate drift via the non-intersection distance, as is done by the drifter R package~\citep{biecek2018dalex}, using 20 bins to calculate distributions. 
This provides the difference between any two given years on a per-covariate basis, which we then average and report in Table~\ref{tab:NRP}.

For example, we may expect some shift in total positive income year to year based on inflation. The non-intersection distance bins any continuous covariates and then provides a distance metric:

\begin{equation}
    d(P,Q) = 1 - \sum_i \min(P_i, Q_i).
\end{equation}

Other distance metrics have been used for such purposes, like the Hellinger Distance or Total Variation Distance~\citep{webb2016characterizing}, but for our purposes NID is adequate to characterize drift.
The year over year drift, as seen in Table~\ref{tab:NRP}, is mostly constant except for 2011-2013, which has a much higher year-over-year average per-covariate drift. 

\section{More Data Details}
\label{app:more_data}
Figure~\ref{fig:IRS_overview} is a figure representing the current audit process at the IRS.

Table~\ref{tab:irs_bandit} is a table of notation correspondence to the IRS equivalent in our structured bandit.

Table~\ref{tab:NRP} is an extended table of summary statistics, including the no change rate in the population NRP sample and the sum of sample weights (equal to the population from which NRP was sampled).

Table~\ref{tab:classcounts} shows the number of taxpayers in each NRP audit class, as well as provides a description of those audit classes. Table~\ref{tab:auditclass_desc} provides a description of those audit classes.

\begin{figure*}[!htbp]
    \centering
    \includegraphics[width=0.9\textwidth]{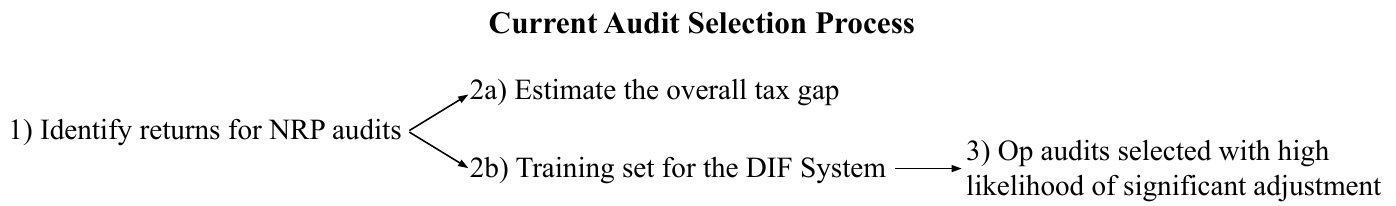}
    \caption{We illustrate the current tax audit process. As described in Section~\ref{sec:background}, the NRP audits are jointly used for population estimation of non-compliance and risk model training. The Op audits are then selected from the DIF model but are never used for estimating the tax gap or for re-training the DIF risk model.}
    \label{fig:IRS_overview}
\end{figure*}

\begin{table}[h!]
\centering
\begin{tabular}{|c|p{4.5cm}|}
\hline
\multicolumn{1}{|l|}{\textbf{Bandit Framework}} & \textbf{IRS Equivalent} \\ \hline
arm ($a_t$) & tax return or taxpayer \\ \hline
context ($X_t^{a}$) & reported information to IRS, in our data 500 covariates constituting mostly of information reported in a tax return \\ \hline
reward ($r_t^a$) & adjustment amount (\$) after audit \\ \hline
timestep ($t$) & the selection year \\ \hline
\end{tabular}%

\caption{Correspondence between structured bandit framework and our setting.}
\label{tab:irs_bandit}
\end{table}

\begin{table*}[!htbp]
    \centering
    \begin{tabular}{ccccccc}
         Year &  \# Audits &  $\mu$-uw & $\mu$-w & Cov. Drift & No Change & $\sum w$\\
         \midrule
         2006 & 13403 & \$2258.07 & \$963.93 & - & 54.8\% & 133M\\
         2007 & 14220 & \$2213.64 & \$920.35 & 0.0143 & 55.1\% & 137M\\
         2008 & 14656 & \$2442.33 & \$938.11 & 0.0141 & 55.7\% & 137M\\
         2009 & 12756 & \$2159.62 & \$989.06 & 0.0197 & 55.5\%&135M\\
         2010 & 13481 & \$2177.26 & \$1034.47 & 0.0143 & 53.8\%&138M\\
         2011 & 13902 & \$3047.39 & \$1038.69& 0.0315 & 53.7\% & 140M\\
         2012 & 15635 & \$3921.93 & \$1041.12& 0.0306 & 49.9\% & 140M\\
         2013 & 14505 & \$3617.64 & \$1173.87& 0.0211 & 49.4\% & 141M\\
         2014 & 14357 & \$5024.25 & \$1218.01& 0.0144 & 47.3\%&143M\\
    \end{tabular}
    \caption{Summary statistics by year of the average misreporting per audited taxpayer across the NRP sample. $\mu$-uw is unweighted mean, $\mu$-w is the mean weighted by NRP sample weights. Cov. drift is the year-over-year covariate drift. No change is the no change rate. $\sum w$ is the sum of the NRP sample weights for a given year, equal to the total sampling population.}
    \label{tab:NRP}
\end{table*}

\begin{table*}

\centering\begin{tabular}
{lrrrrrrrrrrrr}
\toprule
NRP Class &    270 &   271 &    272 &    273 &    274 &    275 &   276 &   277 &    278 &    279 &    280 &    281\\
Year &       &      &       &       &       &       &      &      &       &       &       &       \\
\midrule
2006     &  1972 &  139 &  2311 &  2060 &  1861 &   641 &  427 &  350 &   870 &  1500 &   605 &   667 \\
2007     &  2445 &  138 &  2440 &  1792 &  1762 &   683 &  436 &  414 &   944 &  1680 &   543 &   943 \\
2008     &  2631 &  173 &  2505 &  1797 &  1809 &   658 &  385 &  429 &   943 &  1816 &   603 &   907 \\
2009     &  2688 &  162 &  2143 &  1878 &  1617 &   605 &  335 &  259 &   899 &  1239 &   464 &   467 \\
2010     &  2380 &  163 &  2093 &  1830 &  1722 &   625 &  414 &  349 &  1005 &  1623 &   523 &   754 \\
2011     &  2364 &  164 &  2090 &   965 &  1001 &   761 &  591 &  618 &  1289 &  1589 &   982 &  1488 \\
2012     &  2416 &  214 &  2189 &  1067 &   971 &  1051 &  966 &  962 &   906 &  1726 &  1190 &  1977 \\
2013     &  2540 &  187 &  2211 &  1072 &  1136 &  1002 &  950 &  823 &   914 &  1446 &  1033 &  1191 \\
2014     &  2361 &  215 &  2144 &  1086 &  1067 &   947 &  870 &  832 &   918 &  1449 &   949 &  1519 \\
\bottomrule
\end{tabular}

    \caption{Counts for each NRP class by year in the full NRP sample.}
\label{tab:classcounts}
\end{table*}

\begin{table*}
    \centering
    \begin{tabular}{c|l}
    \toprule
    NRP Activity Code & Description\\
    \midrule
          270 & Form 1040 EITC present \& TPI $<$ \$200k and Sch. C/F Total Gross Receipts (TGR) $<$ \$25k\\
          271 & Form 1040 EITC present \& TPI $<$ \$200,000 and Sch. C/F TGR $>$ \$25,000\\
          272 & Form 1040 TPI $<$ \$200,000 and No Sch. C, E, F or 2106\\ 
          273 & Form 1040 TPI $<$ \$200,000 and No Sch. C or F, but Sch. E or 2106 OKAY\\
          274 & Form 1040 Non-farm Business with Sch. C/F TGR $<$ \$25,000 and TPI $<$ \$200,000\\
          275 & Form 1040 Non-farm Business with Sch. C/F TGR \$25,000 - \$100,000 and TPI $<$ \$200,000\\ 
          276 & Form 1040 Non-farm Business with Sch. C/F TGR \$100,000 - \$200,000 and TPI $<$ \$200,000\\
          277 & Form 1040 Non-farm Business with Sch. C/F TGR \$200,000 or More and TPI $<$ \$200,000\\
          278 & Form 1040 Farm Business Not Classified Elsewhere and TPI $<$ \$200,000 \\
          279 & Form 1040 No Sch. C or F present and TPI $\ge$ \$200,000 and $<$ \$1,000,000 \\
          280 & Form 1040 Sch. C or F present and TPI $\ge$ \$200,000 and $<$ \$1,000,000\\
          281 & Form 1040 TPI $\ge$ \$1,000,000\\
          \bottomrule
    \end{tabular}
    
    \caption{A correspondence of audit classes to their descriptions. Replicated from \protect\url{https://www.irs.gov/pub/irs-pia/2018_doc_6209_section_13.pdf}. TGR is Sch. C/F Total Gross Receipts (TGR). TPI is total positive income. Schedule F is a form filed with tax returns used to report Profit or Loss From Farming. Schedule C is used to report income or loss from a business operated or a profession practiced as a sole proprietor. Schedule E is used to ``report income or loss from rental real estate, royalties, partnerships, S corporations, estates, trusts, and residual interests in real estate mortgage investment conduits (REMICs).'' See \protect\url{https://www.irs.gov/forms-pubs/about-schedule-e-form-1040}. Form 2106 is used to report business expenses. See \protect\url{https://www.irs.gov/forms-pubs/about-form-2106}. EITC is the Earned Income Tax Credit for low-to-moderate-income families and workers. See \protect\url{https://www.irs.gov/credits-deductions/individuals/earned-income-tax-credit-eitc}. }
    \label{tab:auditclass_desc}
\end{table*}

\section{NRP Weights}
\label{app:nrp_weights}

A key complication in our investigation is that the NRP sample used by IRS is a stratified random sample and NRP weights must be used to estimate the population. This makes evaluation on the NRP sample more difficult. As such, we only use NRP weights for the population estimate. However, taking a random sample (as in the $\eps$ exploration sample) means that the sample is not evenly distributed across the population, but rather it matches the distribution of the NRP sample. We weighed alternative designs, such as synthetically replicating features in proportion to NRP weights and then discarding the NRP weights. However, we felt this was not realistic enough. As such, our evaluation might be thought to scale to a larger system in the following way. First a very large NRP sample would be selected. Then, within that larger sample, our methods would select sub-samples that are in line with the true budget.

For all methods we run both a weighted and an unweighted fit using NRP weights for the model fit. We found that ABS was the only method to have reduced variance from an unweighted fit, whereas other methods improved from a weighted fit. This is likely because for non-ABS methods the population estimation mechanism is model-based and the model benefits from having more fine-grained splitting criteria in areas that are up-weighted later on.

The composition of the NRP sample may also add sources of drift and stochasticity to our sample. Each year the NRP sample weights are re-calculated according to changing priorities and improvements to the program. As such, later years have a different composition of samples across NRP activity codes (Table~\ref{tab:auditclass_desc}) than earlier years.

We note that the NRP sample weights are base weights, reflecting the sampling probability, not adjusted for final outcome.

\section{Extended Metrics Descriptions}

\subsection{Choice of cumulative versus average Reward and Total Positive Income}

Note, we report cumulative reward as is standard for bandit settings. Average reward can be recovered by dividing by the number of timesteps. We also note that in early years, where no selection has been made, the selection probability is the same across all algorithms, therefore cumulative reward reflects late-stage differences more clearly.
We report cumulative average TPI for similar reasons. Cumulative average TPI is calculated as the average TPI for a selected batch in a given year, then summed over years.

\subsection{Extended Percent Difference Explanation}
\label{app:pd}
The \textbf{percent difference} is the difference between the estimated population average and the true population average: $100\% * (\hat{\mu} - \mu^*) / \mu^*$. We denote $\sigma_{PE}$ to be the standard deviation of the population estimate percent difference across random seeds. That is, we measure the variation across random seeds on a per year basis, resulting in a standard deviation per year, then we present the average of these standard deviation values. This is in line with current recommendations in the ML community which recommend showing variation across seeds~\citep{henderson2018deep,agarwal2021deep}. This provides insight into the variation of the method across slightly different populations draw from the same distribution. 

$\mu_{PE}$ is absolute mean percent difference across seeds. $\sqrt{|\mu_{PE}|_2^2}$ is the root mean squared percent difference across every prediction (year and seeds). That is, the percent difference for every prediction is squared, averaged, and the square root is taken. This is a scalar point metric and gives some indication as the a combined error rate due to both bias and variance. Note, this takes into account variance inherent to subsampling of NRP (the 80\% sample used to simulate different populations) as well as variance in sampling across seeds. As such, non-model-based methods are at an inherent disadvantage since they do not re-use data from prior years. Though, this may be an interesting direction for future work. 

\subsection{RARE Score}
\label{app:rare}
It can be thought of as a modification of discounted cumulative gain (DCG)~\citep{jarvelin2002cumulated} or RankDCG~\citep{katerenchuk2018rankdcg}. In those methods, the distance between the predicted rank and true rank of a data point is discounted based on the rank position (and in the case of RankDCG, normalized and accounts for ties).
Similar to other related metrics~\citep{jarvelin2002cumulated,katerenchuk2018rankdcg}, the RARE metric takes into account the magnitude of the error in estimate revenue potential as well as the rank. It is effectively the distance from the maximum revenue under the correct ranking, or the percentage of the maximum area under the ranked cumulative reward potential. We consider the true area under the cumulative reward curve where rewards are ordered by magnitude of the true reward of the arm: $\xi_{w, \max} = \sum_{i=0}^N \sum_{j=0}^{i} \sum_{k=0}^{j} w_k r_k$, where $N$ is the size of the total population.

The ranking algorithm's area under the predicted reward curve is denoted by $\hat{\xi}$ and the minimum area under the reward curve is the area under the reverse ordering of the reward $\xi_{\min}$. $w_k$ is the sampling weight in case the training sample is not uniformly drawn (as is the case in NRP). The RARE metric thus gives an approximation of a magnitude-adjusted distance to optimal ranking: $RARE = \frac{\hat{\xi} - \xi_{\min}}{\xi_{\max} - \xi_{\min}}$. After a working group discussion with IRS stakeholders, we found that RARE seemed to capture many key dimensions of interest more than other conventional measures.

\section{Function Approximator Selection}
\label{app:functionapproximator}

We examine which function approximator might be best suited overall if the sample is purely an unbiased random sample. We fit random forests, OLS, and Ridge regression on a purely random sample. We evaluate the ability of the function approximator to rank correctly as well as estimate the population mean. For example, for year 2008, we train on a random sample of 2006 and evaluate the model population estimate and ranking accuracy. For year 2009, we train on a random sample of 2006 and 2007, and so on. 

We find that non-linear estimators consistently achieve significantly higher RARE scores and more accurate population estimates than linear equivalents. As seen in Figure~\ref{fig:function_approximators}. As such, for the remaining experiments, we do not use any linear function approximation methods.

\begin{figure*}[!htbp]
    \centering
    \includegraphics[width=.32\textwidth]{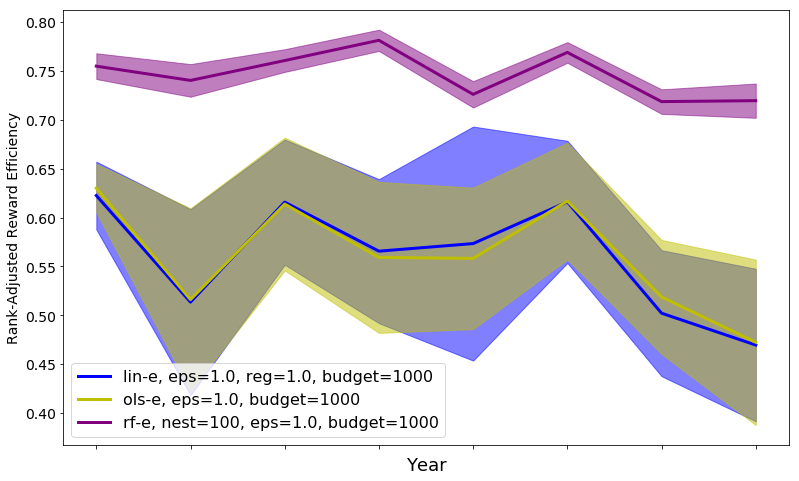}
    \includegraphics[width=.32\textwidth]{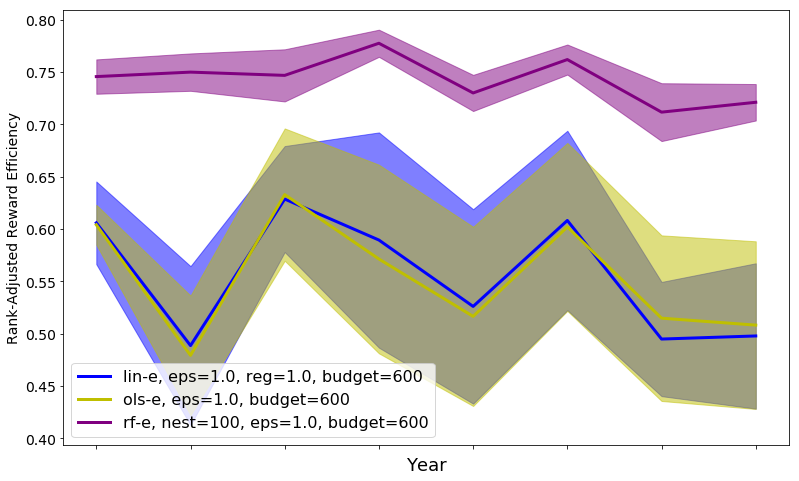}
    \includegraphics[width=.32\textwidth]{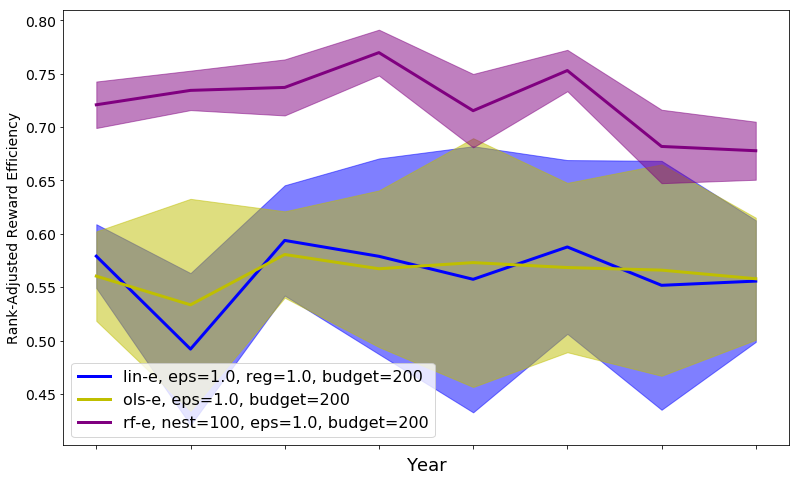}
    \caption{RARE score for linear versus non-linear methods, with non-linear achieving significantly improved performance, especially at higher budgets. ``lin-e'' refers to a linear (ridge regression with regularization factor of 1.0) $\eps$-greedy model, ``ols-e'' to an ordinary least squares (no regularization) $\eps$-greedy model, and ``rf-e'' to a random forest $\eps$-greedy model. Here we plot 30 runs with standard errors. }
    \label{fig:function_approximators}
\end{figure*}

\section{ABS Sampling}
\label{app:abs_sampling} 

In this section we provide further details on ABS, verify that the populate estimate is unbiased, and make some general remarks on the effects of various parameters on the variance of the estimate. Algorithm \ref{alg:abs2} gives an overview of ABS with logistic smoothing. 

Fix a timestep $t$ and let $K$ be our budget. Let $\hat{r}_a=f_{\htheta}(X_t^a)$ be the predicted risk for return $X_t^a$.  First we sample the top $\zeta$ returns. To make the remaining $K-\zeta$ selections, we parameterize the predictions with a logistic function, 

\[\hat{\rho}_a=\frac{1}{1 + \exp(-\alpha (\hat{r}_a-\kappa))},\]

or an exponential function 

\[\hat{\rho}_a = \exp(\alpha \hy_a).\]

$\kappa$ is the value of the $K$-th largest value amongst reward predictions $\{\hat{r}_t^a\}$. For the logistic we normalize such that $\hy_a\in[-5,5]$, and $\hy_a\in[0,1]$ for the exponential. The distribution of transformed predictions $\{\hat{\rho}_a\}$ is then stratified into $H$ non-intersecting strata $S_1,\dots,S_H$. We choose the strata in order to minimize intra-cluster variance, subject to the constraint of having at least $K-\zeta$ points per bin:
\begin{equation}
\label{eq:min_variance1}
    \min_{S_1,\dots,S_H} \quad \sum_{h} \sum_{\hz\in S_h} \norm{\hz - \lambda_h}^2, \quad
    \textnormal{s.t.} \quad |S_h|\geq K-\zeta,
\end{equation}
where $\lambda_h=|S_h|^{-1}\sum_{\hat{\rho}\in S_h}\hz$ is the average value of the points in bin $b$. Note that $\sum_h\sum_{\hz\in S_h}\norm{\hz - \lambda_h}^2=\sum_h\sum_{\hz\in S_h}|S_h|\Var{S_h}$, so the quadratic program \eqref{eq:min_variance} is indeed minimizing intra-cluster variance. 

We place a distribution $(\pi_h)$ over the bins by averaging the risk in each bin, i.e,
\begin{equation}
\label{eq:pi1}
    \pi_h = \frac{\lambda_h}{\sum_{h'}\lambda_{h'}}.
\end{equation}
To make our selection, we sample $K-\zeta$ times from $(\pi_h,\dots,\pi_H)$ to obtain a bin, and then we sample \emph{uniformly} within that bin to choose the return. We do not recalculate $(\pi_1,\dots,\pi_H)$ after each selection, so while we are sampling without replacement at the level of returns (we cannot audit the same taxpayer twice), we are sampling with replacement at the level of bins. This is (i) because of computational feasibility, and (ii) in order to obtain an unbiased estimate of the mean via HT Sampling~\cite{horvitz1952generalization}.

In particular, note that the probability that arm $a$ in stratum $S_h$ is sampled is $p_a=(K-\zeta)\pi_h/N_h$ (see the next subsection for a derivation), where $N_h=|S_h|$ is the size of $S_h$. 

\begin{theorem}
If $\sel$ is the set of returns chosen for auditing and $S_{H+1}$ contains those $\zeta$ points first sampled, then
\begin{equation*}
    \hmuht(t) = \frac{1}{\sum_a w_a}\bigg(\sum_{a\in \sel\setminus S_{H+1}} \frac{w_ar_a}{p_a} + \sum_{a\in S_{H+1}}w_ar_a\bigg),
\end{equation*}
is an unbiased estimator of the true mean 
\begin{equation*}
    \mu(t) = \frac{1}{\sum_a w_a}\sum_{a}w_ar_a.
\end{equation*}
\end{theorem}

\begin{proof}
To see this, let $\ind_{a\in\sel}$ be the random variable indicating whether arm $a$ is sampled.  Since $\E[\ind_{a\in\sel}]=p_a$, linearity of expectation gives
\begin{align*}
    \E[\hmuht]&=\frac{1}{\sum_a w_a}\E\bigg[\sum_{a\in\mathcal{A}\setminus S_{H+1}}\frac{w_ar_a}{p_a}\ind_{a\in \sel} +\sum_{a\in S_{H+1}}w_ar_a\bigg] \\
    &= \frac{1}{\sum_a w_a}\bigg(\sum_{a\in\mathcal{A}\setminus S_{H+1}}\frac{w_ar_a}{p_a}\E[\ind_{a\in \sel}] +\sum_{a\in S_{H+1}}w_ar_a\bigg)\\
    &= \frac{1}{\sum_a w_a}\sum_{a\in \mathcal{A}} w_ar_a.
\end{align*}
\end{proof}

\begin{algorithm}[tb]
   \caption{Adaptive Bin Sampling - Logistic Smoothing}
   \label{alg:abs2}
\begin{algorithmic}
   \STATE {\bfseries Input:} $\alpha$, $H$, $\zeta$, $K$, $(X_0, r_0)$
   \STATE Train model $f_{\htheta}$ on initial data $(X_0,r_0)$. 
   \FOR{$t=1,\dots,T$}
   \STATE Receive observations $X_t$
   \STATE Predict rewards $\hy_a = f_{\htheta}(x_a)$. 
   \STATE Sample top $\zeta$ predictions.
   \STATE For all $a$ compute $\hat{\rho}_a \gets (1 + \exp(-\alpha(\hy_a - \kappa))^{-1}$ 
   \STATE Construct strata $S_1,\dots,S_H$ by solving \eqref{eq:min_variance}. 
   \STATE Form distribution $\{\pi_h\}$ over strata via \eqref{eq:pi}. 
    \REPEAT  
   \STATE $h\sim (\pi_1,\dots,\pi_H)$ 
   \STATE Sample return at uniformly at random from $S_h$.
   \UNTIL{$K-\zeta$ samples drawn}
   \STATE Compute $\hmuht$ once true rewards are collected. 
   \STATE Retrain model $\hat{f}$ on $(\cup_i^t X_i, \cup_i^t r_i)$.  
   \ENDFOR
\end{algorithmic}
\end{algorithm}

\subsection{Variance of Population Estimate}
Write the HT estimator as 
\begin{equation*}
    \hmuht = \frac{1}{N}\sum_a \frac{r_a}{p_a}\ind_{a\in \sel},
\end{equation*}
where $\sel$ is the set of selected arms and $p_a=\Pr(a\in \sel)$ is arm $a$'s inclusion probability in $\sel$. Then
\begin{align*}
    \Var(\hmuht) &= \frac{1}{N^2}\sum_{a,b}\frac{r_ar_b}{p_ap_b}\Cov(\ind_{a\in \sel}\ind_{b\in \sel}) \\
    &= \frac{1}{N^2}\bigg(\sum_a \frac{r_a^2}{p_a}(1-p_a) + \sum_a\sum_{b\neq a}\frac{r_ar_b}{p_ap_b}(p_{a,b}-p_ap_b)\bigg),
\end{align*}
where $p_{a,b}=\Pr(a,b\in \sel)=p_{b,a}$ is the joint inclusion probability of arms $a$ and $b$. Note that for the $\zeta$ arms in stratum $S_{h+1}$, $p_a=1$ and $p_{a,b}=p_b$. Therefore, all terms involving such arms are zero and they do not contribute to the variance.  

We can make this expression more specific to our case by rewriting the inclusion probabilities as functions of the strata. Fix an arm $a$ and suppose it's in stratum $S_h$. Let $m=K-\zeta$ be the number of returns we're randomly sampling (i.e., discarding those $\zeta$ points greedily chosen from the top of the risk distribution). The law of total probability over the $m$ trials gives 
\begin{equation*}
    p_a = \sum_{\ell=0}^m \Pr(a\in \sel||S_h\cap \sel|=\ell)\Pr(|S_h\cap \sel|=\ell).
\end{equation*}
The first term in the product is the probability that $a$ is chosen as one of $\ell$ elements in a bucket of size $N_h=|S_h|$. The second term is the probability that $S_h$ was selected precisely $\ell$ times and is distributed as a binomial. Therefore, 
\begin{align*}
    p_a &= \sum_{\ell=0}^m \frac{\ell}{N_h}{K\choose \ell} \pi_h^\ell(1-\pi_h)^{m-\ell} = \frac{m\pi_h}{N_h}.
\end{align*}
Now consider $p_{a,b}$ for distinct arms $a,b$. Let $b\in S_g$. Conditioning on $b\in \sel$ gives $\Pr(a\in\sel|b\in\sel)=\frac{(m-1)\pi_h}{N_h}$ if $g\neq h$ since there are now $m-1$ trials to select $a$. If $g=h$, then $\Pr(a\in\sel|b\in\sel)=\frac{(K-1)\pi_h}{N_h-1}$ since there are $m-1$ trials to select $a$ from a bin of size $N_h-1$. 
Thus 
\begin{align*}
  p_{a,b} &=\Pr(a\in\sel|b\in\sel)\Pr(b\in\sel)\\
  &=\begin{cases}\frac{m(m-1)\pi_h\pi_g}{N_hN_g},&\text{if }g\neq h,\\
\frac{m(m-1)\pi_h^2}{N_h(N_h-1)},&\text{if }g=h.
\end{cases}  
\end{align*}
Rewriting the variance as a summation over the strata, we see that the variance is the difference of two terms $V_1$ and $V_2$ where $V_1$ as a sum across all strata and $V_2$ includes cross-terms dependent on the relationship between strata.
\begin{equation*}
    \Var(\hmuht) = \frac{1}{mN^2}(V_1 - V_2),
\end{equation*}
where 
\begin{equation*}
    V_1 = \sum_{h=1}^H (N_h\pi_h^{-1}-m)\sum_{a\in S_h}r_a^2,
\end{equation*}
and 
\begin{equation*}
V_2 =  \sum_{h=1}^H\sum_{a\in S_h}r_a
     \bigg(\frac{N_h-m}{N_h-1}\sum_{b\in S_h\setminus a}r_b + \sum_{g\neq h}\sum_{b\in S_g}r_b\bigg)   .
\end{equation*}
We make a few remarks on the variance here, but leave a full analysis to future work. If the budget is small relative to the strata sizes (as is the case here), then $\frac{N_h-m}{N_h-1}\approx 1$, and $V_2$ reduces to $\sum_a\sum_{b\neq a}r_ar_b$ which is independent of the strata. As $\alpha$ grows and we place more weight on those returns deemed higher risk by the model, $p_a\to0$ for lower risk arms. This results in many arms clustered in a few strata with high $N_h$ and low $\pi_h$, which increases $V_1$. Also, as $\zeta$ grows and we perform more greedy sampling, $m$ decreases and the variance increases roughly proportionally.

\section{Experimental Setup}
\label{app:expsetup2}
Figure~\ref{fig:processdiagram} provides a visual aid to help understand the problem setting.

\begin{figure}[H]
    \centering
    \includegraphics[width=.5\textwidth]{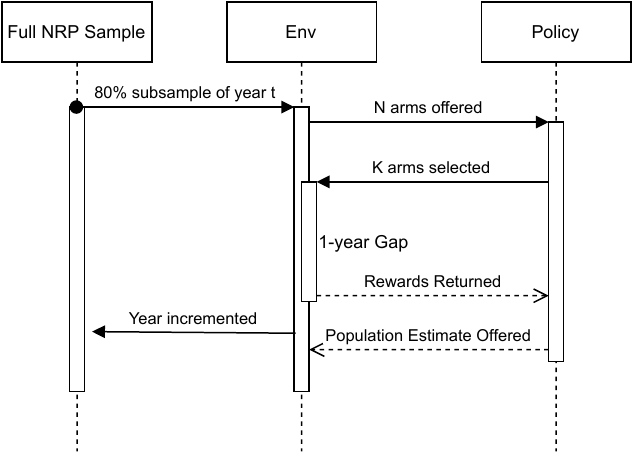}
    \caption{Diagram describing our experimental setup.}
    \label{fig:processdiagram}
\end{figure}

\section{Hyperparameter Tuning}
\label{app:hyperpararmeters}

The ideal approach would tune the hyperparameters for function approximators using cross-validation every time the model is fit in the active learning process. 
However, we found this approach to be extremely computationally expensive, a small grid of experiments requiring over a week to run. 
In the interest of reducing energy consumption -- see, for example, discussion by \citet{henderson2020towards} -- we instead opt for a less computationally expensive proxy.
We take train our function approximators on 2006 and then evaluate their RARE score and population estimate for the 2008 year, using 5-fold cross-validation across both years. We then run a grid search for all function approximators used. Finally we find the top point on the smoothed Pareto frontier between RARE and population estimation to find the optimal hyperparameters. To do this we rank the reward and population estimation criteria based on 
Since there is concept drift from year to year, we expect that these hyperparameters are sub-optimal and results may be even further improved with careful per-year hyperparameter tuning. However, this approach is sufficient for the purposes of our experiments. 

For handling hyperparameters of the sampling algorithms, we rely on sensitivity analyses rather than hyperparameter searches. This is in line with recent work that promotes reporting results over ranges of hyperparameters and random seeds, particularly for sequential decision-making systems~\citep{henderson2018deep,bouthillier2021accounting,agarwal2021deep}. Hyperparameter grids were run as follows in Table~\ref{tab:hparamgrid}.

\begin{table}[!htbp]
    \centering
    \begin{tabular}{c|c}
    \toprule
    Method & Hyperparameter Grid\\
    \midrule
       $\eps$-Greedy  & $\eps=0.0, 0.1, 0.2, 0.4$ \\
       & weighted, unweighted fit\\
       \hline
       UCB  & $Z=0.1, 1.0, 10.0, 100.0$ \\
       &weighted, unweighted fit\\
       \hline
       ABS & mixing=exponential, logistic \\
       & $\alpha=0.1, 0.5, 1.0, 1.5, 2.0, 5, 10, 15$ \\ 
       & $\zeta=0.0,0.2,0.4, 0.6, 0.8$ \\
       & trim=0\%, 2.5\%, 5\% \\
       & weighted, unweighted fit\\
       \bottomrule
    \end{tabular}
    \caption{Hyperparameter grids}
    \label{tab:hparamgrid}
\end{table}

For Table~\ref{tab:best_reward}, the only place where we display single hyperparameter settings, we use the following selection protocol from the grid of hyperparameters above. 
Given the lack of longitudinal data, we rely on intra-year data sub-sampling for validation sets. We select 5 random seeds, corresponding to different validation subsampled datasets. 
First, we identify the top reward band by finding methods with overlapping confidence intervals on reward. See \citet{joseph2018meritocratic} for discussion on meritocratic fairness given overlapping confidence intervals. 
Then, we order results by root mean squared error and select the top hyperaparameters for each method according to these validation seeds. We then use results from all train seeds to get results displayed in the paper.

\section{Winsorization}

We winsorize the rewards returned such that the top 1\% of highest values is set to the top 99th percentile's value. Negative values were truncated to 0. This is in line with recommendations for existing models within the IRS~\citep{latenonfiler} and other research on audit data~\citep{debacker2015importing}. This helps to stabilize predictions, protecting against unusually large outlier adjustments, but may bias the model. 

\section{Budget Selection}
\label{app:budgets}

At each timestep the IRS can select a limited budget of samples -- for example the NRP sample in 2014 was 14,357 audits. This is a tiny fraction of audits as compared to the general population of taxpayers -- and thus impossible to replicate when using the NRP sample to evaluate selection mechanisms. The goal of the NRP sample is to select a large enough sample to approximate the taxpayer base. The parallel in our experiments would be to ensure we select a sample which is smaller than the coreset needed to model the entire data. 

Another way of thinking about the size of the budget allowed per year to approximate the NRP mechanism is by determining what is the minimum random sample to achieve a 3\% margin of error with 95\% confidence of the NRP sample population, per the 2018 OMB requirements for IPERIA reporting. Using an 80\%  (from subsampling) sample of the per-year average of 14102 NRP yearly samples, we are left with an average of 11282 arms per year. We should need about a 975 arm budget for a random sampling mechanism (ignoring stratification) to achieve OMB specifications.

We then use the approach of \citet{sener2018active} to find a minimal coreset which a model could use to achieve a reasonable fit.
We first fit a random forest to the entire dataset for a given year and calculate the residuals (or the mean squared error across the dataset).  We then iteratively select batches of 25 samples according to the method presented by \citet{sener2018active}.
We use only the raw features to compute distance (contrasting the embedding space used by the authors).
We refit a random forest with the same hyperparameters as the optimal fit on the smaller coreset sampler. Then we calculate the ratio of the mean squared error on the entire year's data to the optimal mean squared error.
We find that the mean squared error is reduced to roughly 2x the overfit model at around 600 coreset samples, reducing very slowly after that point.
We find that around 600 samples is the absolute minimum number required to reduce the mean squared error to a stable level at 2 times the optimal mean squared error. To simulate the small sample sizes of the NRP selection, we select this smaller budget of 600 as our main evaluation budget size, corresponding to roughly 4\% of the 2014 NRP sample.

Note that in practice, IRS faces heterogeneous costs for audits. For the purposes of this work, we assume a fixed budget of arm/tax returns rather than a fixed budget of auditor hours.

\section{Confidence Intervals on the Time Series}
\label{app:confidenceintervals}

Since arms within a year are randomly sampled, this is close to the subsampled bootstrap mechanism used for time-series as described by \citet{politis1994large,politisromanoj,politis2003impact}. We consider the year-by-year NRP sample as a stationary time series with each year of arms as an identically distributed sample, though year-to-year the samples are not necessarily independent. The subsample boostrap provides a mechanism to estimate confidence intervals for such a time series (approximately). This is also similar to the delete-$d$ where $d = n - b$ jackknife as described in \citep{politisromanoj,shao1989general}. Because of the computational complexity of running experiments on the time-series, we keep $b$ low at 20 bootstrap samples with distinct random seeds (and thus the setting is not identical to the delete-$d$ jackknife).

\section{Results}
\label{app:allresults}

First, we provide an extended graph with hyperparameter details for the main settings presented in Table~\ref{tab:best_reward}. ABS-1 is a hyperparameter configuration that focuses slightly more on reward at the cost of population estimation variance. $\epsilon$-Greedy uses an $\epsilon$ of 0.1, UCB-1 has $Z=1$, UCB-2 has a larger exploration factor of $Z=10$. ABS-1 uses an exponential mixing function with 80\% greedy sample, $\alpha=5$, and a 2.5\% trim factor. ABS-2 uses a logistic mixing function, $\alpha=0.5$, a 5\% trim, and 80\% greedy sample. Both ABS methods use an unweighted fit while all other approaches saw improved results with a weighted fit.

\begin{table*}[!htbp]
    \centering
    {\bf Best Reward Settings}\\
    \begin{tabular}{r|l|l|l|l|l|l}
    \hline 
\emph{Policy} & $\mu_{NR}$ & $\mu_{PE}$ & $\sqrt{|\mu_{PE}|_2^2}$ & $\mu_{RARE}$ & ${R}$ & $\sigma_R$\\
\hline 
Greedy & 36.5\% & 16.4 & 21.5 & 0.70 & \$43.6M* & \$760k\\
UCB-1 & 38.6\%  & 15.3 & 20.9 & 0.70 & \$42.4M* & \$853k\\
ABS-1 & 37.6\%  & 0.4 & 34.2 & 0.70 & \$41.5M* & \$796k\\
$\epsilon$-Greedy & 38.3\%  & 6.1 & 10.5 &  0.73 & \$41.3M* & \$772k\\
UCB-2 & 40.7\%  & 15.6 & 22.2 & 0.70 & \$40.7M* & \$1.2M\\
ABS-2 & 38.3\%  & 0.6 & 26.8 & 0.71 & \$40.5M* & \$672k\\
Random & 53.1\%  & 1.5 & 13.1 & - & \$12.7M & \$493k\\
\end{tabular}
    \caption{We rank all methods and hyperparameters based on reward bands. We show the top reward band and best hyperparameter settings for each method in that reward band (where CIs across random seeds overlap). $R$ is the average cumulative reward at the final timestep across random seeds, $\mu_{NR}$ is the average no-change rate, $\mu_{RARE}$ the average RARE score, $\mu_{PE}$ the absolute percent difference of the population estimate, and  $\sigma_{PE}$ the standard deviation of the population estimate.  $\sqrt{|\mu_{PE}|_2^2}$ is root mean squared error.}
    \label{tab:best_reward_app}
\end{table*}

\subsection{$\eps$-Greedy, Random Sample-Only for Population Estimation}
\label{app:eps}

It may be tempting to use only the random sample of the $\eps$-Greedy methods for population estimation, but we note that in the constrained budget setting we investigate here the variance of these estimates becomes much higher than ABS settings with comparable rewards. This demonstrates the utility of re-using information to navigate the bias-variance-reward trade-off problem. For a budget of 600, for example, Table~\ref{tab:eps-only} demonstrates this. Though the relationship is somewhat non-linear, ABS always has the potential for pareto improvement, yielding higher reward for the same variance. For example, though ABS-2 has 4 more standard deviations higher it yields over \$7M extra in revenue. And though ABS-1 yields similar rewards to $\eps=0.1$, it is 6.4 standard deviations lower in variance.

\begin{table*}[!htbp]
    \centering
    \begin{tabular}{c|c|c|c}
    \toprule
         $\eps$& $\sigma_{PE}$ & $\sqrt{|\mu_{PE}|_2^2}$ & $R$ \\
         \midrule
         0.1 & 37.4 & 40.8 & \$41.3M \\
         0.2 & 29.9 & 32.2 & \$38.7M \\
         0.4 & 20.6 & 22.04 & \$33.1M \\
         \midrule
         ABS-1 & 31.0 & 34.2 & \$41.5M\\
         ABS-2 & 24.5 & 26.8 & \$40.5M\\
         \bottomrule
    \end{tabular}
    \caption{The population estimation variance across random seeds and reward given different amounts of $\eps$-Greedy exploration samples.}
    \label{tab:eps-only}
\end{table*}

\subsection{Price-for-variance Trade-offs and Policy Implications}
\label{app:pricemetric}

Using our sampling mechanism we can roughly fit a function to roughly approximate the Pareto front which helps understand the ``price-for-variance'' trade-off. First, we take the Pareto front from ABS sampled hyperparameters in Figure~\ref{fig:variance}. Then, we fit a cubic to this Pareto front as seen in Figure~\ref{fig:marginal}.

\begin{figure}[H]
    \centering
    \includegraphics[width=.45\textwidth]{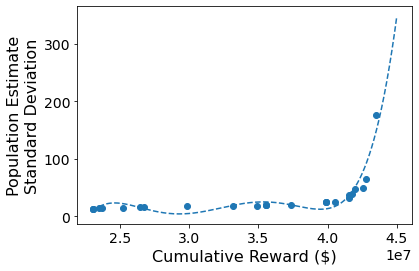}
    \caption{The pareto front of ABS hyperparameters with a cubic fit. This can be thought of as the marginal benefit for giving up reward.}
    \label{fig:marginal}
\end{figure}

Note this should be interpreted with caution since different algorithms may have different Pareto fronts and population drift may change things. Additionally fitting polynomials or exponential to this front can have its own inference challenges. But this gives a rough understanding that under ABS (and even under $\epsilon$-Greedy which has a slightly worse navigation of this pareto frontier), it is worth giving up some reward initially to cut variance in half, but beyond a certain point it is not worth the trade-off (at least for ABS which is able to navigate the frontier more closely.) Note the fit above yields this polynomial.

\begin{align*}
    \mathbb{V} &= 2.11026680 \times 10^{-33} r^5\\ &-3.39482296 \times 10^{-25} r^4 \\&+ 2.16187688 \times 10^{-17} r^3 \\ &-6.80927492\times10^{-10} r^2 \\&+ 
        1.06045458\times 10^{-2} r\\& -6.53025073\times 10^4
\end{align*}

Since ABS allows for some empirical examination of these trade-offs, this provides some information for policymakers to understand what are the trade-offs for increasing the amount of random samples. For example, when OMB specifies some guidelines for estimation, our experimental exploration here provides a foundation for future work in examining this reward-variance trade-off and navigating policy implications.

\subsection{Using RFR significantly outperforms LDA and incorporating non-random data helps}
\label{app:lda}

For the LDA baseline rather than regressing a predicted reward value for selection, we predict whether the true reward is $>$\$200 (our no-change cutoff). Arms are selected based on increasing likelihood that they are part of the \$200+ reward class. This is included both as context to our broad modeling decisions, and as an imperfect stylized proxy for one component of the current risk-based selection approach used by the IRS.

Figure~\ref{fig:lda} shows that the cumulative return of $\epsilon$-greedy sampling strategies using RFR-based approaches are significantly higher than LDA-based approaches.
We emphasize again that the LDA model we use here is a \emph{stylized} approximation of the current risk-based selection process and does not incorporate other policy objectives and confidential mechanisms of DIF.
We note that while it serves as a \emph{rough} baseline model, it demonstrates again that regression-based and globally non-linear models are needed for optimal performance in complex administrative settings such as this.

Fitting to \emph{only} the random data, even with the RFR approach, reduces the ability of the model to make comparable selections and increases the variance across selection strategies. This can be seen in Figure~\ref{fig:lda}, where fitting to random only leads to between \$6.2M and \$1.3M less reward cumulatively (95\% CIs on effect size) and standard deviation across seeds is increased by \$700k. Since the current risk-selection approach mainly uses the NRP model, this suggests that future work on incorporating Op audits into the model training mechanism could improve overall risk-selection. This of course bears the risk of exacerbating biases and should be done with careful correction for data imbalances.

In Figure~\ref{fig:lda} we show where LDA fits in terms of reward maximization, achieving significantly less reward than RF-based methods.

\begin{figure}
\centering
    \includegraphics[width=0.6\textwidth]{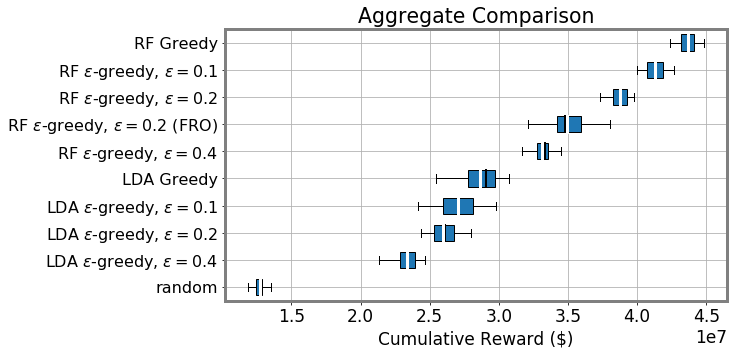}

    \caption{A comparison of cumulative reward earned between LDA-based approaches and RFR-based approaches. FRO indicates that the RFR was fit to the random sample only.}
    \label{fig:lda}
\end{figure}

\subsection{Fine-grained Kernel Density}

Figure~\ref{fig:samplingdist2} shows the kernel density plots for every year in the time series on one random seed.

\begin{figure*}[!htbp]
    \centering
    \includegraphics[width=.23\textwidth]{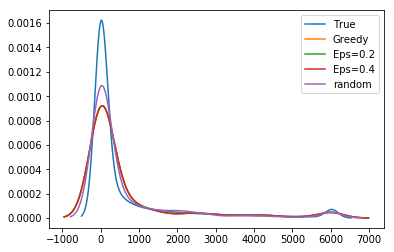}
    \includegraphics[width=.23\textwidth]{images/2007.png}
    \includegraphics[width=.23\textwidth]{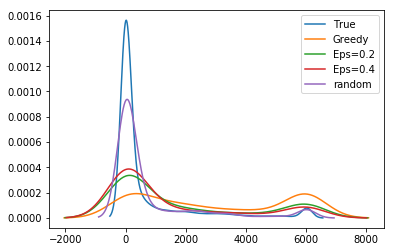}
    \includegraphics[width=.23\textwidth]{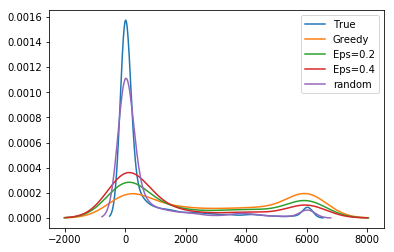}
    \includegraphics[width=.23\textwidth]{images/2010.png}
    \includegraphics[width=.23\textwidth]{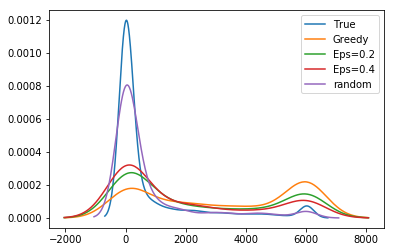}
    \includegraphics[width=.23\textwidth]{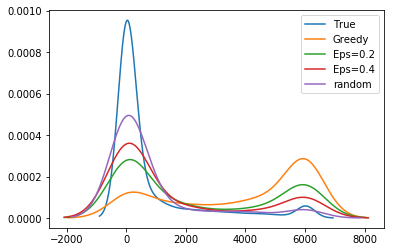}
    \includegraphics[width=.23\textwidth]{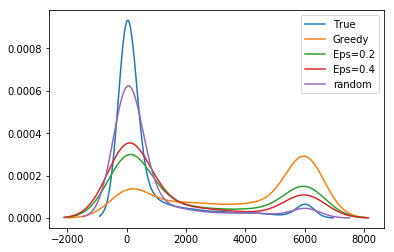}
        \includegraphics[width=.23\textwidth]{images/2014.png}

    \caption{A kernel density plot of the distribution of sampled arms from 2006 (top left) to 2014 (bottom). X-axis is true reward. Y-axis is sampling distribution density.}
    \label{fig:samplingdist2}
\end{figure*}

\subsection{Ablation of ABS}
Here, we plot the effect of various hyperparameters for each model. Figures \ref{fig:abs_results} illustrate the effect of model parameters on cumulative reward, population estimation (mean), population estimation (variance), no change rate, and RARE score. For each parameter and each metric, we plot the results of all runs with the same parameter as it varies across its range. We use violin plots to demonstrate the density of the values.    

\begin{figure*}[!htbp]
    \begin{center}
        \textbf{ABS}
    \end{center}
    \centering
    \includegraphics[scale=0.2]{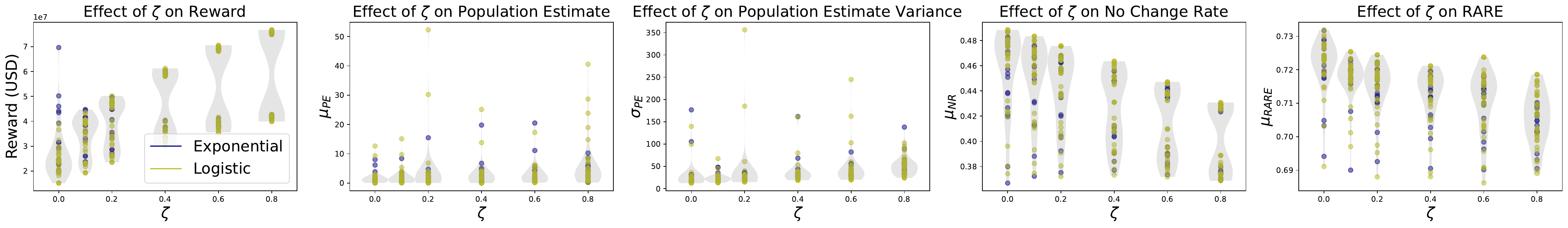}
    \includegraphics[scale=0.2]{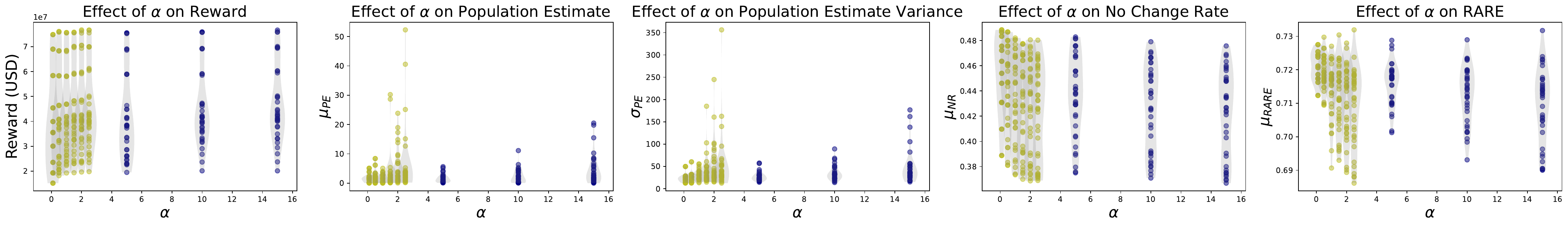}
    \includegraphics[scale=0.2]{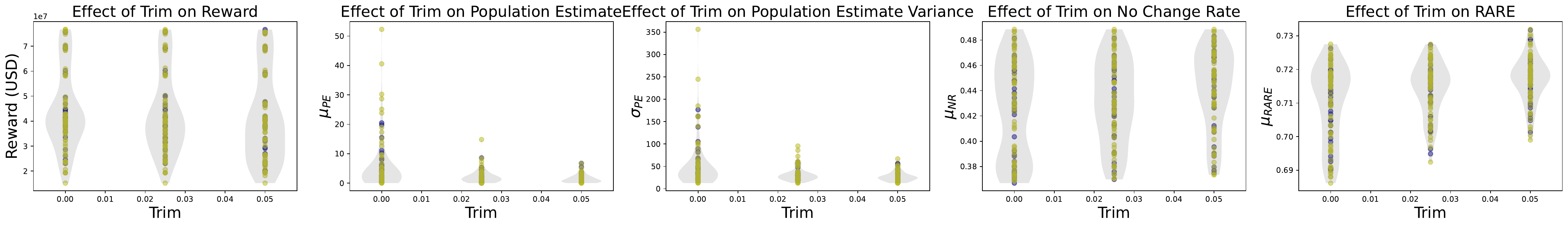}
    \caption{Results for ABS with a budget of 600 arms. The top row explores the effect of $\zeta$ on five metrics, the middle row explores the effect of $\alpha$, and the bottom the effect of the trimming factor.}
    \label{fig:abs_results}
\end{figure*}

\subsection{Larger Budget}

Results stay similar if we increase the budget to 1000 arms per timestep. This can be seen in Figures~\ref{fig:abs_results1000} and~\ref{fig:more_results1000}.

\begin{figure*}[!htbp]
    \begin{center}
        \textbf{ABS}
    \end{center}
    \centering
    \includegraphics[scale=0.2]{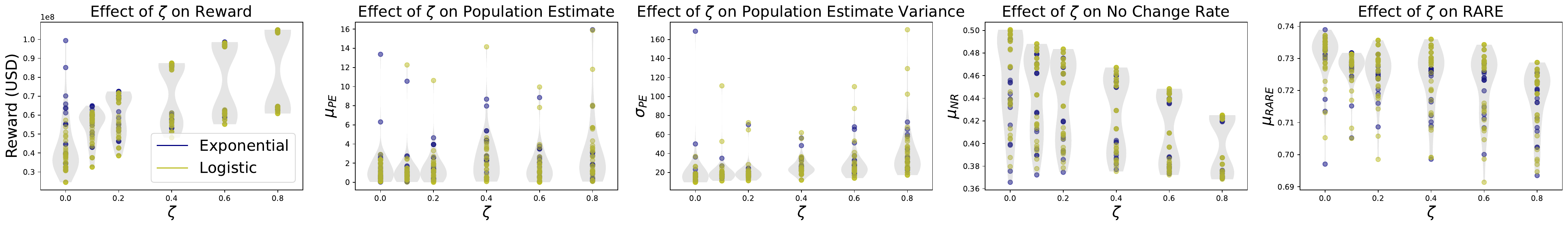}
    \includegraphics[scale=0.2]{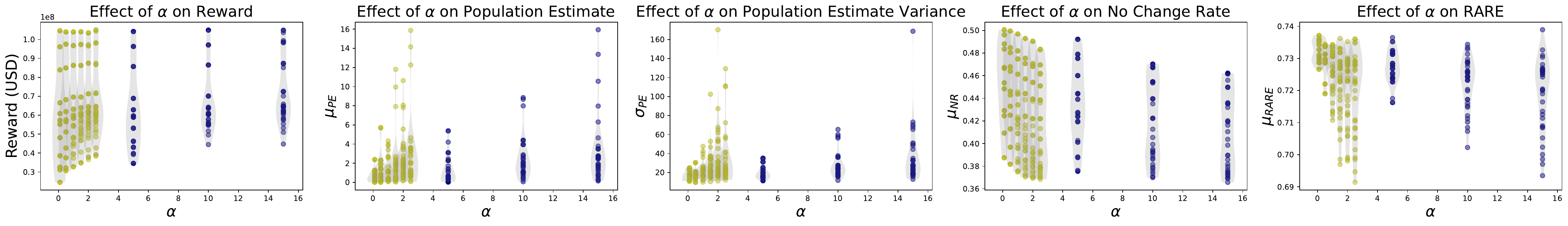}
    \includegraphics[scale=0.2]{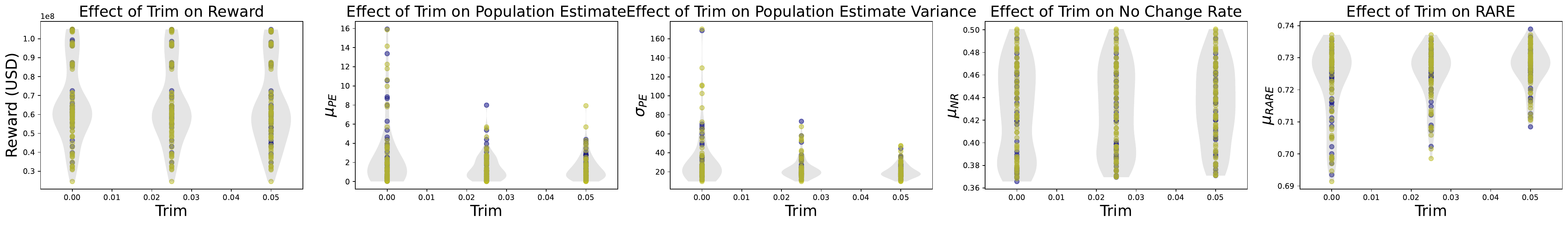}
    \caption{Results for ABS with a budget of 1000 arms. The top row explores the effect of $\zeta$ (Z) on five metrics, the middle row explores the effect of $\alpha$, and the bottom the effect of the trimming factor.}
    \label{fig:abs_results1000}
\end{figure*}
\begin{figure*}[!htbp]
    \centering
    \includegraphics[width=.48\textwidth]{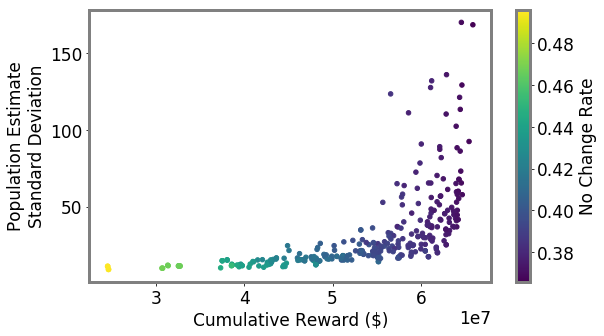}
    \includegraphics[width=.48\textwidth]{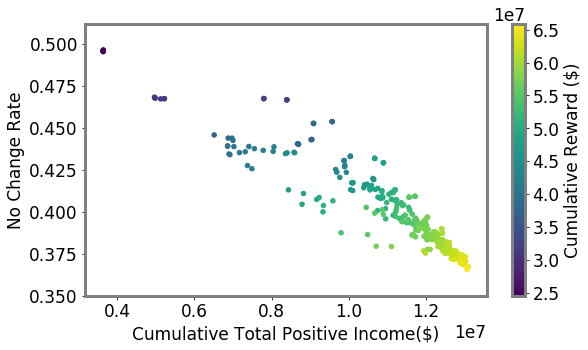}
    \includegraphics[width=.48\textwidth]{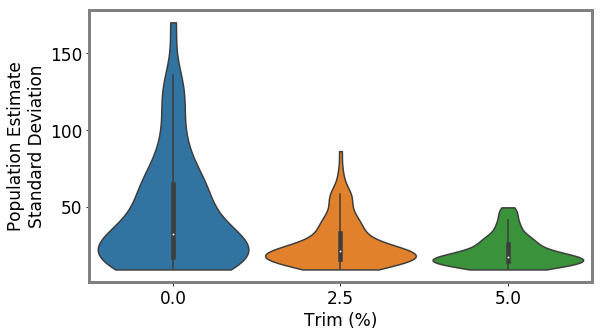}
    \includegraphics[width=.48\textwidth]{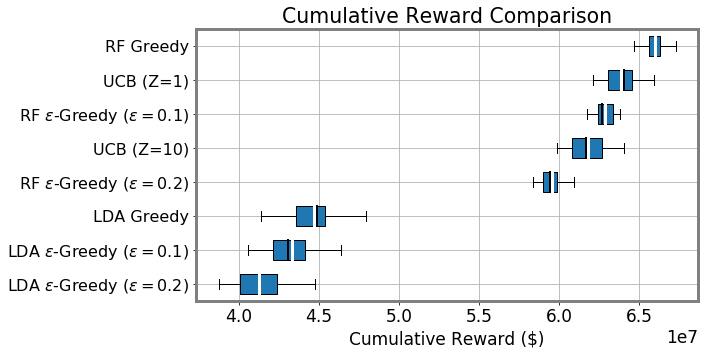}
    \caption{Top Left, Middle, Right: ABS hyperparameters with a budget of 1000 arms. Trends largely reflect results at the budget of 600 arms. Bottom: similarly, cumulative rewards for greedy and UCB methods follow trends as in the 600 budget setting.}
    \label{fig:more_results1000}
\end{figure*}

\subsection{Sampling breakdown by classes}

Figure~\ref{fig:moreclass} breaks down ABS and other methods in terms of audit classes.

\begin{figure*}[!htbp]

\centering
\includegraphics[width=\textwidth]{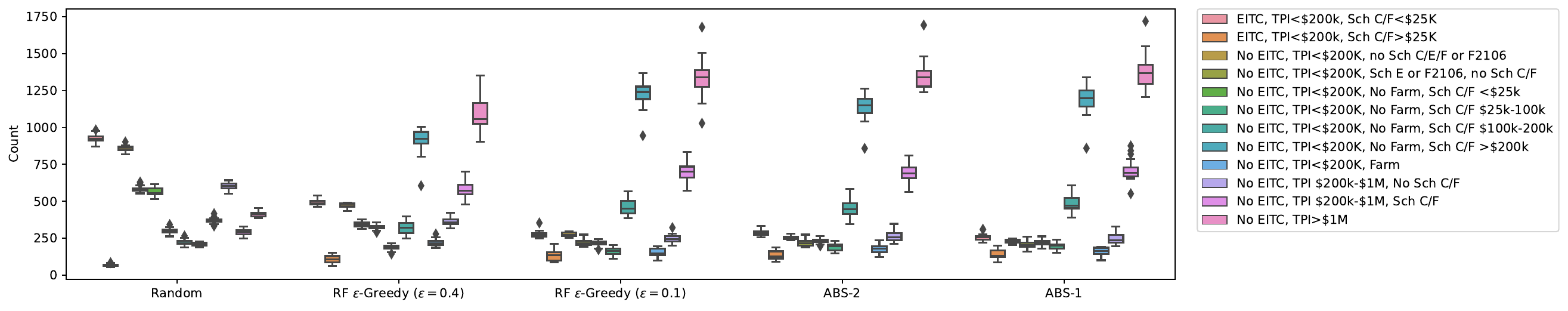}
\includegraphics[width=\textwidth]{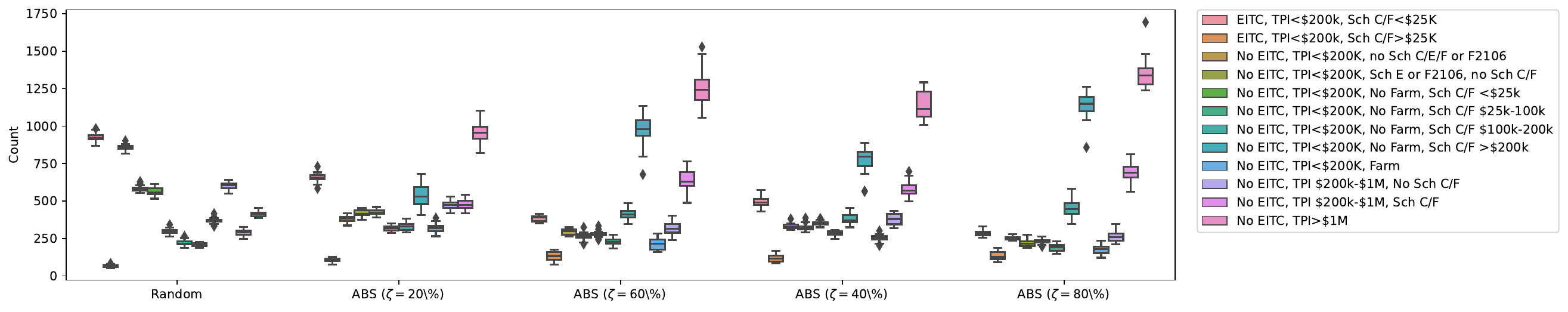}

\caption{Top: Comparison of the top ABS hyperparameter settings seen in Table~\ref{tab:best_reward} with RF $\epsilon$-greedy as a reference. Bottom: A small ablation of $\zeta$'s effect on class selection counts for an ABS setting of logistic mixing function, $\alpha=0.5$ and a 5\% trim. }
\label{fig:moreclass}
\end{figure*}

\subsection{Expanded Figure}

To ensure that both a log scale and non-log scale version of Figure~\ref{fig:variance} are available, we include both of those in Figure~\ref{fig:exp2fig}.

\begin{figure*}[!htbp]

\centering
\includegraphics[width=.65\textwidth]{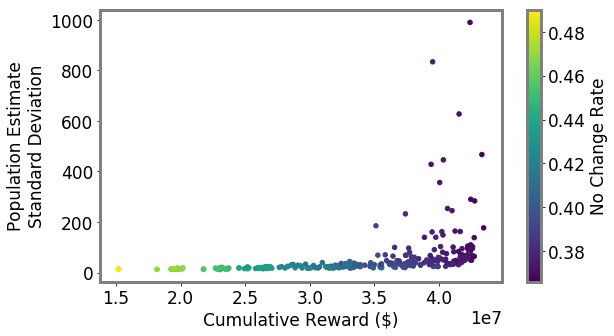}
\includegraphics[width=.65\textwidth]{images/figure2.pdf}

\caption{For clarity we include a non-log scale version of Figure~\ref{fig:variance} results and compare it with the other version.}
\label{fig:exp2fig}
\end{figure*}

\newcommand{\ben}[1]{{\blue} #1\xspace} 

\newcommand{\pred}{\widehat{r}}

\renewcommand{\H}{\mathcal{H}}
\newcommand{\A}{\mathcal{A}}
\newcommand{\hatA}{\widehat{\mathcal{A}}}
\newcommand{\reg}{\mathsf{Reg}}
\newcommand{\hreg}{\widehat{\reg}}
\newcommand{\x}{\vb{x}}
\newcommand{\hx}{\hat{x}}
\newcommand{\D}{\mathcal{D}}
\newcommand{\F}{\mathcal{F}}

\renewcommand{\H}{\mathcal{H}}
\renewcommand{\a}{\vb{a}}
\newcommand{\ha}{\hat{a}}
\newcommand{\Rel}{\mathsc{Rel}}
\newcommand{\Relhat}{\widehat{\mathsc{Rel}}}
\newcommand{\reals}{\mathbb{R}}
\newcommand{\e}{\boldsymbol{e}}
\def\EE{{\mathbb{E}}}\def\PP{{\mathbb{P}}}
\newcommand{\RegEmp}{\widehat{\mathsf{R}}}
\newcommand{\xtm}{{\sf eXtreme}\xspace}
\newcommand{\igw}{{\sf IGW}\xspace}
\newcommand{\absv}{{\sf ABS}}
\newcommand{\bse}{{\sf beam}\xspace}
\newcommand{\chosenset}{\mathcal{A}}
\newcommand{\feedbackset}{\Phi}
\newcommand\numberthis{\addtocounter{equation}{1}\tag{\theequation}}

\section{Regret}
\label{app:regret}

Proving regret bounds in the non-linear batched optimize-and-estimate structured bandit poses a significant challenge and sets the ground for exciting new theoretical research. Nonetheless, we identify an initial mechanism to provide some level of bounded regret for ABS. We ground our work upon that of \citet{sen2021top} and \citet{simchi2020bypassing}. In the setting of \citet{sen2021top} there are an extremely large, but finite, number of arms. During each epoch, the agent is presented with a single context ($x$), drawn i.i.d. from a distribution $\mathcal{D}$. Using this context the agent must select $K$ arms. \citet{sen2021top} use an Inverse Gap Weighting (IGW) strategy to select these $K$ arms. 
IGW was introduced by~\citep{abe1999associative} and has been leveraged to prove regret bounds in realizable settings with general function classes~\citep{foster2020beyond,simchi2020bypassing,foster2020instance,sen2021top,zhu2022contextual}. 
In this setting, 
given a set of arms $\cA$, 
an estimate $\pred:\cX\times\cA\to\reals$ of the reward function, 
and a context $x$, 
there is a distribution over arms assigned: $p= \text{IGW}\left(\cA; \pred(x, \cdot)\right)$ 
\begin{align*}
    p(a | x) = \begin{cases} 
        \frac{1}{|A| + \gamma_\tau(\pred(x, a_{\star}) - \pred(x, a))} & \mbox{if } a \neq a_{\star} \\
        1 - \sum_{(N-\zeta+D) \in \cA: (N-\zeta+D) \neq a_{\star}} p((N-\zeta+D)|x) & \mbox{otherwise} 
    \end{cases}
\end{align*}
where $a_{\star} = \argmax_{a \in \cA}\pred(x, a)$, $\gamma_\tau$ is a scaling factor, and where $A$ is the total number of arms in an epoch $\tau$.  In the algorithm proposed by \citet{sen2021top}, for a $k$-armed bandit the top $\zeta$ arms are chosen greedily while $K-\zeta$ arms are chosen according to this probability distribution.

Crucially, and why the IGW algorithm does not apply to the optimize-and-estimate setting, the IGW algorithm used in that case cannot be trivially used for unbiased population estimates. To select $K$ arms, the bandit incrementally recalculates probabilities after selecting each arm. This dynamic re-calculation makes ensuring unbiasedness difficult, because determining the true (or even approximate) inclusion probabilities quickly becomes non-trivial. ABS instead leverages stratification to keep exact probabilities and ensure unbiasedness.

Conveniently, however, we can reduce the ABS and optimize-and-estimate setting to a variant of the top-$k$ contextual bandit and IGW algorithm. This allows us to leverage the tools of \citet{sen2021top} and \citet{simchi2020bypassing} to prove bounded regret for a particular instantiation of ABS. We note that we demonstrate effectiveness in other settings empirically, as others like \citet{sen2021top} do.

Note that in the $k$-armed contextual bandit setting it is assumed that there is one context per epoch as opposed to a per-arm context ($x_a$). To make the proof tractable, we assume that arms are not volatile and that the context vector contains taxpayer information for all arms. That is, we reduce the optimize-and-estimate structured bandit to a contextual bandit with some additional constraints. 
Thus we assume a fixed number of arms available year over year ($N$) and a fixed budget ($K$) over the time period ending in timestep $T$.
We assume for convenience that the context is drawn i.i.d. from the distribution of possible populations that could be drawn year over year. That is, we have some global distribution $\D$ and every timestep, $N$ observations $x_1,\dots,x_N\sim\D$ are drawn i.i.d from $\D$. And $x_t$ can be thought of as a vector representing information about all taxpayers, to simplify the proof. The goal is then to estimate the population conditional on the draw. This of course, is a simplification of the true setting, and we note future theoretical work can tackle the more complicated dependent population draw.

Assume for the sake of our reduction proof, that we too, like \citet{sen2021top} update in epochs. That is, we have an epoch $\tau$ at which we have updated our regression model. For $n_{\tau}$ steps we then re-use this regression model year over year. This is not too far off from reality. Since we receive delayed rewards, we would not update our model for several years anyways, aligning with the epoch structure. Note, then that, aligning with the notation of \citet{sen2021top}, we have epochs $\tau = 1, ..., \hat{\tau}(T)$. Thus,  $t_{\hat{\tau}(T)} = \sum_{\tau=\tau}^{\hat{\tau}(T)} n_{\tau} = T$ and timestep $t$ is simply $\sum_{i=1}^{\tau} n_{i}$ where $n_\tau$ is the number of steps between updating the model at epoch $\tau$. 

Effectively, we decompose our problem into accumulation steps before doing a regression update, like our reward delay does in reality. Assume that our functional structure is such that all non-informative parts of the context (taxpayer information not indexed by $a$) is masked out. Thus for action $a$ we denote the context as $x_a$ indicating a masked and indexed portion of the full context. For notational convenience, we let $\hat{\A}$ denote both the selection algorithm itself (i.e., ABS)  as well as the set of selected actions in a given step. That is, we will write $a\in\hat{\A_t}$ to denote the arm sampled by $\hat{\A}$ at time $t$ and $x_a$ would be the context associated with that arm (e.g., tax return covariates for a selected filer). If $t$ is clear from context, we drop it from the notation for clarity.

Next, we continue the reduction of ABS to the IGW setting to leverage the tools required to prove regret bounds. Recall that ABS has the distribution $p = \absv \left( \cA ; \pred(x, \cdot) \right)$, for the selection probability of one exploratory arm (we only prove this case, keeping in line with \citet{sen2021top}):

\begin{align*}
    p(a | x) = \frac{1}{|S_h|}\frac{|S_h|^{-1}\sum_{a \in S_h} g(\pred(x,a))}{\sum_{S_h \in S} |S_h|^{-1}\sum_{a \in S_h} g(\pred(x,a))},
    \numberthis
\label{eq:abs_mixture}
\end{align*}

where $g$ is a function class mapping the risk/reward distribution that is monotone (order preserving). And noticing that uniform selection of one arm (for the single exploratory arm case) is uniform over the selected stratum. Assume now that for the purposes of this analysis $g(y)$ is the function:

\begin{align*}
    g(\pred) = \frac{1}{N-\zeta + \gamma_\tau (\pred(x,a^*) - \pred_\tau(x,a))},
    \numberthis
\label{eq:functionalform}
\end{align*}

where again $a^{\star} = \argmax_{a \in \cA}\pred(x, a)$, $\gamma_\tau$ is a scaling factor, and where $A$ is the total number of arms available at epoch $\tau$. This is, trivially, a monotonic function that fits into the ABS framework since, as \citet{sen2021top} do, we assume all $\pred(x,a) \in [0,1]$.

\begin{lemma}[ABS Minimum Probability Bound]
\label{lem:abs_bound}
For ABS at epoch $\tau$, we have
\begin{equation}
    p(a|x) \geq \frac{(N-\zeta)}{HN_S}\bigg(\frac{1}{N-\zeta + \gamma_\tau(\pred(x, a^*)-\pred_\tau(x, a)+d_S)}\bigg),
\end{equation}

where $N_S=\max_h\{|S_h|\}$ is the maximum number of observations in any stratum, $d_S$ is the maximum distance between rewards in a bin, and H is the number of strata.

\end{lemma}

\begin{proof} From the definition of an ABS mixture function in Eq.~\ref{eq:abs_mixture} and our chosen functional form in Eq.~\ref{eq:functionalform}:
\begin{align*}
    p(a | x) = \frac{1}{|S_h|}\frac{|S_h|^{-1} \sum_{a \in S_h} \frac{1}{N-\zeta + \gamma_\tau (\pred(x,a^*) - \pred(x,a))}}{\sum_{S_h \in S} |S_h|^{-1} \sum_{a \in S_h} \frac{1}{N-\zeta + \gamma_\tau (\pred(x,a^*) - \pred(x,a))}},
\end{align*}

and re-arranging terms and settings $Z=\sum_{S_h \in S} |S_h|^{-1} \sum_{a \in S_h} \frac{1}{N-\zeta + \gamma_\tau (\pred(x,a^*) - \pred(x,a))}$

\begin{align*}
    p(a | x) = \frac{1}{|S_h|} \frac{1}{Z} |S_h|^{-1} \sum_{a \in S_h} \left[ \frac{1}{\left(N-\zeta + \gamma_\tau  (\pred(x,a^*) - \pred(x,a)) \right)} \right].
\end{align*}

Consider that if we assume that the maximal distance for any pair of points in any cluster is $d_S$ then this value is:
\begin{align*}
    \ge \frac{1}{|S_h|} \frac{1}{Z} \left[\frac{1}{\left(N-\zeta + \gamma_\tau  (\pred(x,a^*) - (\pred(x,a)-d_S) \right)} \right].
\end{align*}

and assuming that $r \in [0,1]$ and $\pred \in [0,1]$ then $Z$ at its largest (and hence the probability at its smallest) is $Z \le \frac{H}{N-\zeta}$. This is because the number of clusters is $H$ each with a maximal average of $N-\zeta$ if $\pred(x,a^*) - \pred(x,a) \rightarrow 0$. 

Thus

\begin{align*}
    Z&=\sum_{S_h \in S} |S_h|^{-1} \sum_{a \in S_h} \frac{1}{N-\zeta + \gamma_\tau (\pred(x,a^*) - \pred(x,a))}\\
    &\le \sum_{S_h \in S} |S_h|^{-1} \sum_{a \in S_h} \frac{1}{N-\zeta}\\
    &\le \sum_{S_h \in S} \frac{1}{N-\zeta}\\
    &\le H\frac{1}{N-\zeta}\\
\end{align*}

Putting it together we have:

\begin{align*}
    p(a|x) \geq \frac{1}{|S_h|} \frac{N-\zeta}{H}\bigg(\frac{1}{N-\zeta + \gamma_\tau(\pred(x, a^*)-\pred_\tau(x, a)+d_S)}\bigg).
\end{align*}

For $N_S=\max_h \{|S_h|\}$, this bound resolves to 

\begin{align*}
    p(a|x) \geq  \frac{N-\zeta}{H\cdot N_S}\bigg(\frac{1}{N-\zeta + \gamma_\tau(\pred(x, a^*)-\pred_\tau(x, a)+d_S)}\bigg).
\end{align*}

\end{proof}

Note that as the number of strata goes to $N-\zeta$ and the intrastratum variation shrinks, the probability converges to the top-$k$ bandit, but provable unbiased population estimates become difficult.

Under this functional form for the mixture function, we can bound that regret as follows. As is the case for \citet{sen2021top}, we examine only the case for $K-\zeta=1$ exploratory arm. Nonetheless, our experiments focus on proxies that nonetheless perform well, as \citet{sen2021top} do in their setting as well. 
Note that it is likely that, unlike for the IGW method, our ABS regret bound may hold for $K-\zeta>1$. This is because, again, we do not need to re-calculate probabilities throughout an epoch, but rather they are fixed throughout the entirety of the epoch. However, we leave this as a challenge for future theoretical research in this area.

In order to prove our main theorem, we assume that the true reward mapping is a member of the family of estimators. This is Assumption 1 from \citet{sen2021top}.

\begin{assumption} [{\bf Realizability}]
	\label{asum:realizability}  
	At every time $t$, 
	there exists an $\hat{r} \in \mathcal{F}$ such for all $a\in\A_t$,
	\begin{align}
	  \E_{x\sim\D}[r^*(x,a)] = \hat{r}(x, a), 
	\end{align}
where $\cF$ is the class of reward functions.  %
\end{assumption}

As in \citet{sen2021top}, we assume that $\cF$ is a finite class of functions.

Note that \citet{sen2021top} make an assumption on partial feedback (their Assumption 3) such that there is a factor $c$ that governs that minimum probability that you will get feedback for a given selected arm. For our case $c=1$ since we always receive feedback. Thus, we remove $c$ constants by setting them to 1 throughout.

\begin{theorem}
\label{thm:topkregret} As in \citet{sen2021top}, under the previously stated assumptions, and assuming function form $g(\pred)$ in Eq.~\ref{eq:functionalform}, when run with parameters
\begin{align*}
    &r = 1;~~~~ t_{\tau} = 2^\tau; ~~~~ \gamma_\tau = \frac{1}{32}\sqrt{\frac{(N-\zeta + D)t_{\tau-1}}{162\log \left( \frac{|\cF| T^3}{\delta}\right)}}, ~~~~ d_S = D/\gamma_\tau, ~~~~ N_S = (N-\zeta)/H
\end{align*}
has regret bound
\begin{align*}
    R(T) = \cO \left(\log(T) + K\sqrt{(N-\zeta + D) T  \log \left( \frac{|\cF| T}{\delta} \right)} \right)
\end{align*}
with probability at least $1 - \delta$, for a finite function class $\cF$ where $d_S$ is the maximal distance between predicted reward in any ABS stratum and $D$ is a chosen constraint, for simplicity we also assume that strata have equal numbers of arms $N_S$, that there is one exploratory arm ($r=1$), and that the number of steps between reward function updates is $d^\tau$ where $\tau$ is an epoch.
\end{theorem}

Notice that both regret bounds increase with the width of the strata in ABS. So keeping the strata constrained (as with the nearest neighbor clustering we employ in practice), keeps the regret bound constrained as well. We note that it is likely possible to give tighter PAC bounds for strata under some other assumptions, but we leave this to future work as the focus of our work is not theoretical but rather the introduction of the optimize-and-estimate structured bandit setting and the IRS study. Finally, we note that there is some minimal strata size that is required to preserve unbiasedness. We cannot use the method of \citet{sen2021top} directly for the optimize-and-estimate structured setting because we require unbiasedness guarantees. ABS provides the required changes to remain unbiased while minimizing regret.

To prove this regret bound, we closely follow the proof structure of \citet{sen2021top}. For clarity, when portions of the proof structure do not change, we will refer and reproduce their Lemmas without reproducing full proofs. Where we must make modifications for our setting we provide the full modification here.

Let $K$ be the budget. Let $\H_{t-1}$ be the natural filtration at time $t-1$.

Thus, we can borrow the following combined Lemma from \citet[supplemental page 3,10-12]{sen2021top},

\begin{lemma}
\label{lem:square_prob_bound}
For any $\delta<1/e$, the event 
\begin{equation*}
    \mathcal{E} = \bigg\{\tau \ge 2: \frac{2}{t_{\tau-1}} \sum_{s=1}^{t_{\tau-1}} \E_{x \sim \D,\hatA_s}\bigg[\frac{1}{K}\sum_{a\in\hatA_s}(r^*(x, a)-\pred_\tau(x, a))^2|\H_{s-1}\bigg] \le \phi_\tau^2\bigg\},
\end{equation*}
holds with probability at least $1-\delta$. Where the event holds if 
\begin{equation*}
    \phi_\tau = \sqrt{81 \log \left(\frac{|\Fcal|t_{\tau-1}^3}{\delta}\right)}
\end{equation*}
under the realizability assumption and otherwise holds under the $\epsilon$-realizability assumption if 

\begin{equation*}
    \phi_\tau = \sqrt{210 \log \left(\frac{|\Fcal|t_{\tau-1}^3}{\delta}\right)}
\end{equation*}
\end{lemma}

\begin{proof}
See \citet[supplemental page 3,10-12]{sen2021top} for proofs.
\end{proof}

\begin{lemma}
\label{lem:blorg}
Fix epoch $\tau$. Consider the set $\{a_1,\dots,a_k\}$ of arms selected by an arbitrary policy such that arms are unique (non-overlapping) in the set. Let $G\subset \A_t$ be the set of $\zeta$ observations selected greedily by $\A$. If event in Eq.~\ref{lem:square_prob_bound} holds, then: 
\begin{align*}
   \E_{x\sim\D} \frac{1}{K}\sum_{i=1}^K \Delta_{i,\tau}         &\leq \phi_\tau \bigg( (N-\zeta + D)  + \frac{1}{t_{\tau-1}}\sum_{t} \gamma_{\hat{\tau}(t)} \E_{x\sim\D}\frac{1}{K} \sum_i (\pred_\tau(x, \hat{a}_i)) - \pred_\tau(x, a_i)\bigg)^{1/2} \\ 
\end{align*}
\end{lemma}
\begin{proof}
For the sake of brevity, let $\Delta_{i,\tau} = |\pred_\tau(x, a_i) - r^*(x, a_i)|$. 
Notice that since all observations are drawn iid from $\D$ which does not change, we can average over timesteps and write 
\begin{align*}
    \E_{x \sim \D}\bigg[\sum_{i=1}^K \Delta_{i,\tau}\bigg] = \frac{1}{t_{\tau-1}}\sum_{t=1}^{t_{\tau-1}} \E_{x \sim \D}\bigg[\sum_{i=1}^K \Delta_{i,\tau} \bigg].  
\end{align*}

Recalling that $t_{\tau}$ is all timesteps up to a given epoch $\tau$. 
Splitting the sum into those observations in $G$ and those not in $G$, we get 
\begin{align*}
    \E_{x\sim\D} \sum_{i=1}^K \Delta_{i,\tau} &= \frac{1}{t_{\tau-1}}\sum_t \E_{x\sim\D}\bigg[\frac{1}{K}\sum_{i=1}^K \Delta_{i,\tau}\ind(a_i\in G) + \frac{1}{K}\sum_{i=1}^K \Delta_{i,\tau}\ind(a_i\notin G)\bigg] \\ 
    &= \frac{1}{t_{\tau-1}}\sum_t\E_{x\sim\D} \frac{1}{K} \sum_{i=1}^K \Delta_{i,\tau}\ind(a_i\in G) + \frac{1}{t_{\tau-1}}\sum_t\E_{x\sim\D} \frac{1}{K}\sum_{i=1}^K \Delta_{i,\tau}\sqrt{\frac{p_{\hat{\tau}(t)}(x, a_i)}{p_{\hat{\tau}(t)}(x, a_i)}}\ind(a_i\notin G). 
\end{align*}
Note that the probabilities $p_{\hat{\tau}(t)}(x, a_i)$ are functions of the epoch $\tau$ and for any individual timestep we can map back to the epoch using $\hat{\tau}(t)$.

Recall that by Cauchy-Schwarz you have:
\begin{align*}
     \bigg[\sum_i^n z_i \bigg] \leq |n|\bigg(\sum_i z_i^2\bigg)^{\frac{1}{2}}.
\end{align*}

Apply this as follows:
\begin{align*}
    \frac{1}{t_{\tau-1}}\sum_t\E_{x\sim\D}\frac{1}{K}\sum_{i=1}^K \Delta_{i,\tau} \ind(a_i\in G) \leq \bigg(\bigg(\frac{1}{t_{\tau-1}}\sum_t\E_{x\sim\D}\frac{1}{K}\sum_{i=1}^K \Delta_{i,\tau} \ind(a_i\in G)\bigg)^2 \bigg)^{1/2}. 
\end{align*}

Trivially moving the square in by one:
\begin{align*}
    \frac{1}{t_{\tau-1}}\sum_t\E_{x\sim\D}\frac{1}{K}\sum_{i=1}^K \Delta_{i,\tau} \ind(a_i\in G) \leq \bigg(\frac{1}{t_{\tau-1}^2}\bigg(\sum_t\E_{x\sim\D}\frac{1}{K}\sum_{i=1}^K \Delta_{i,\tau} \ind(a_i\in G)\bigg)^2 \bigg)^{1/2}. 
\end{align*}

Then by Cauchy-Schwarz and trivially moving the square inside so that we have $(\frac{1}{K})^2$:
\begin{align*}
    \frac{1}{t_{\tau-1}}\sum_t\E_{x\sim\D}\frac{1}{K}\sum_{i=1}^K \Delta_{i,\tau} \ind(a_i\in G) \leq \bigg(\frac{t_{\tau-1}}{t_{\tau-1}^2}\sum_t\E_{x\sim\D}\frac{1}{K^2}\bigg(\sum_{i=1}^K \Delta_{i,\tau} \ind(a_i\in G)\bigg)^2 \bigg)^{1/2}. 
\end{align*}

Applying Cauchy-Schwarz again:
\begin{align*}
    \frac{1}{t_{\tau-1}}\sum_t\E_{x\sim\D}\frac{1}{K}\sum_{i=1}^K \Delta_{i,\tau} \ind(a_i\in G) \leq \bigg(\frac{1}{t_{\tau-1}}\sum_t\E_{x\sim\D}\frac{K}{K^2}\sum_{i=1}^K \bigg(\Delta_{i,\tau} \ind(a_i\in G)\bigg)^2 \bigg)^{1/2}. 
\end{align*}

Which reduces to:

\begin{align*}
    \frac{1}{t_{\tau-1}}\sum_t\E_{x\sim\D}\frac{1}{K}\sum_{i=1}^K \Delta_{i,\tau} \ind(a_i\in G) \leq \bigg(\frac{1}{t_{\tau-1}}\sum_t\E_{x\sim\D}\frac{1}{K}\sum_{i=1}^K \Delta_{i,\tau}^2\ind(a_i\in G)\bigg)^{1/2}. 
\end{align*}

For the second, Cauchy-Schwarz gives:
\begin{align*}
    \frac{1}{t_{\tau-1}}\sum_t\E_{x\sim\D}\frac{1}{K}\sum_{i=1}^K \Delta_{i,\tau}\sqrt{\frac{p(x, a_i)}{p(x, a_i)}}\ind(x, a_i\notin G) &\leq \bigg(\frac{1}{t_{\tau-1}}\sum_t\E_{x\sim\D}\frac{1}{K}\sum_{i=1}^K \Delta_{i,\tau}^2 p_{\hat{\tau}(t)}(x, a_i)\bigg)^{1/2}\\ & \hspace{1cm} \times \bigg(\frac{1}{t_{\tau-1}}\sum_t\E_{x\sim\D}\frac{1}{K}\sum_{i=1}^Kp_{\hat{\tau}(t)}(x, a_i)^{-1}\ind(a_i\notin G)\bigg)^{1/2} 
\end{align*}
Combining the previous two inequalities, and noticing $\sum_i\Delta_{i,\tau}^2\ind(a_i\in G)\leq \sum_{a_i\in G}\Delta_{i,\tau}^2$,
\begin{align*}
    \E_{x\sim\D}\frac{1}{K}\sum_{i=1}^K \Delta_{i,\tau} &\leq  \bigg(\frac{1}{t_{\tau-1}}\sum_t\E_{x\sim\D}\frac{1}{K} \sum_{a_i\in G}^K \Delta_{i,\tau}^2\bigg)^{1/2} + \bigg(\frac{1}{t_{\tau-1}}\sum_t \E_{x\sim\D}\frac{1}{K}\sum_{i=1}^K \Delta_{i,\tau}^2 p_{\hat{\tau}(t)}(x, a_i)\bigg)^{1/2}\\& \hspace{2.25in} \times \bigg(\frac{1}{t_{\tau-1}}\sum_t\E_{x\sim\D}\frac{1}{K}\sum_{i=1}^Kp_{\hat{\tau}(t)}(x, a_i)^{-1}\ind(a_i\notin G)\bigg)^{1/2}.
\end{align*}

Notice that for any realization of $\D$, $|G|<K$ by construction, so the rightmost sum has at least one non-zero term. Moreover, since $p(a_i)\leq 1$, $p(a_i)^{-1}\geq 1$, so this sum is at least 1. It follows that  
\begin{align*}
    \E_{x\sim\D} \frac{1}{K}\sum_{i=1}^K \Delta_{i,\tau} &\leq  \bigg\{\bigg(\frac{1}{t_{\tau-1}}\sum_t\E_{x\sim\D} \frac{1}{K} \sum_{a_i\in G} \Delta_{i,\tau}^2\bigg)^{1/2} + \bigg(\frac{1}{t_{\tau-1}}\sum_t \E_{x\sim\D}\frac{1}{K}\sum_{i=1}^K \Delta_{i,\tau}^2 p_{\hat{\tau}(t)}(x, a_i)\bigg)^{1/2}\bigg\}\\
    &\hspace{2.25in}\times \bigg(\frac{1}{t_{\tau-1}}\sum_t\E_{x\sim\D}\frac{1}{K}\sum_{i=1}^Kp_{\hat{\tau}(t)}(x, a_i)^{-1}\ind(a_i\notin G)\bigg)^{1/2} \\ 
    &\leq \bigg(\frac{2}{t_{\tau-1}}\sum_t\E_{x\sim\D}\bigg(\frac{1}{K}\sum_{a_i\in G}^K \Delta_{i,\tau}^2 + \frac{1}{K}\sum_{i=1}^K \Delta_{i,\tau}^2 p_{\hat{\tau}(t)}(x, a_i)\bigg)\bigg)^{1/2}\\
    &\hspace{2in}\times \bigg(\frac{1}{t_{\tau-1}}\sum_t\E_{x\sim\D}\frac{1}{K}\sum_{i=1}^Kp_{\hat{\tau}(t)}(x, a_i)^{-1}\ind(a_i\notin G)\bigg)^{1/2} \\ 
    &\leq \bigg(\frac{2}{t_{\tau-1}}\sum_t \frac{1}{K} \E_{x\sim\D}\bigg(\sum_{a_i\in G} \Delta_{i,\tau}^2 + \E_{a_i \sim p_{\hat{\tau}(t)}} \Delta_{i,\tau}^2 p_{\hat{\tau}(t)}(x, a_i)\bigg)\bigg)^{1/2}\\
    &\hspace{2in}\times \bigg(\frac{1}{t_{\tau-1}}\sum_t\E_{x\sim\D}\frac{1}{K}\sum_{i=1}^Kp_{\hat{\tau}(t)}(x, a_i)^{-1}\ind(a_i\notin G)\bigg)^{1/2} \\ 
        &\leq \bigg(\frac{2}{t_{\tau-1}}\sum_t \frac{1}{K} \E_{x\sim\D} \E_{\hatA_t} \sum_{a_i \in \hatA_t} \Delta_{i,\tau}^2\bigg)^{1/2}\\
    &\hspace{2in}\times \bigg(\frac{1}{t_{\tau-1}}\sum_t\E_{x\sim\D}\frac{1}{K}\sum_{i=1}^Kp_{\hat{\tau}(t)}(x, a_i)^{-1}\ind(a_i\notin G)\bigg)^{1/2} \\ 
    \numberthis
    \label{eq:twoparthing}
\end{align*}
where the second inequality follows since  $\sqrt{a} + \sqrt{b}\leq (2(a+b))^{1/2}$ for $a,b\geq 0$.

Now, by Lemma~\ref{lem:abs_bound} and noticing that we only sample from $N-\zeta$ arms after greedy arms are chosen, so our population for sampling shrinks. 
For this case $\hat{a}$ is the $i$-th arm that is selected under the predictive policy as compared to a general strategy set $a_i$. 

\begin{align*}
  \frac{1}{p_{\hat{\tau}(t)}(x, a_i)}\ind(a_i \notin G) &\leq  \frac{HN_S}{N-\zeta}( (N-\zeta) + \gamma_\tau(\pred_\tau(x, \hat{a}_k) - \pred_\tau(x, a_i) + d_S)) 
\end{align*}

For simplicity, let us assume that each bin has a uniform number of arms $N_S = \frac{N-\zeta}{H}$.
We note that this creates a very loose downstream regret bound, but helps us avoid complications from an extra multiplicative term. Since these regret bounds are not a core function of this work, future work might tighten the regret bound. We have 

\begin{align*}
\frac{1}{p_{\hat{\tau}(t)}(x, a_i)}\ind(a_i \notin G) &\le (N-\zeta) +  \gamma_\tau(\pred_\tau(x, \hat{a}_k) - \pred_\tau(x, a_i) + d_S).\\
 &\le (N-\zeta) +  \gamma_\tau(\pred_\tau(x, \hat{a}_k) - \pred_\tau(x, a_i)) + \gamma_\tau d_S.
 \end{align*}
 
 Define $D=d_S\gamma_\tau$. Then, 
 
 \begin{align*}
 \frac{1}{p_{\hat{\tau}(t)}(x, a_i)}\ind(a_i \notin G) &\le (N-\zeta + D) + \gamma_\tau(\pred_\tau(x, \hat{a}_k)) - \pred_\tau(x, a_i)).
\end{align*}

Then by Lemma 3 from \citet[supplemental p.5]{sen2021top} (noting the change from $k$-index to $i$ index for the top value):

 \begin{align*}
 \frac{1}{K} \sum_i^K \frac{1}{p_{\hat{\tau}(t)}(x, a_i)}\ind(a_i \notin G) &\le (N-\zeta + D)  + \frac{1}{K} \sum_i \gamma_\tau(\pred_\tau(x, \hat{a}_i)) - \pred_\tau(x, a_i)).
\end{align*}

Recall Lemma~\ref{lem:square_prob_bound}, using this for the first term of Eq.~\ref{eq:twoparthing} (see also \citet[suppelemental p.5]{sen2021top}), we can move $\phi_\tau$ into the equation and create a new bound:

\begin{align*}
   \E_{x\sim\D} \frac{1}{K}\sum_{i=1}^K \Delta_{i,\tau}         &\leq \phi_\tau \bigg( (N-\zeta + D)  + \frac{1}{t_{\tau-1}}\sum_{t} \gamma_{\hat{\tau}(t)} \E_{x\sim\D}\frac{1}{K} \sum_i (\pred_\tau(x, \hat{a}_i)) - \pred_\tau(x, a_i)\bigg)^{1/2} \\ 
\end{align*}
from which the result follows. 

\end{proof}

Now we modify the induction hypothesis of \citet{sen2021top} such that we take into account the width of the ABS strata. First, recall the induction hypothesis from \citet{sen2021top} (modified for ABS) is as follows. 

\paragraph{Induction hypothesis ($\tau$):} For any epoch $m<\tau$, let $a_i^*$ denote the action with the $i$-th highest reward. Let $\ha_i$ denote the observation with the \emph{estimated} $i$-th highest reward, according to the model $\pred$. For any set of $K$ selected actions $\a=(a_1,\dots,a_k) \in \hat{\cA}$, let 
\[\reg(\a) = \E_x \sum_{i=1}^k [r^*(x,a_i^*) - r^*(x,a_i)],\]
be the regret of selecting the actions $\a$, and 
\[\hreg_m(\a)=\E_x\sum_{i=1}^k [\pred_m(x,\ha_i) - \pred_m(x,a_i)],\]
the estimated regret according the $\pred_m$ where $m$ is the current epoch and the model fit to that point. The following result bounds the difference between the true regret and estimated regret.

For any set of $K$ distinct actions $(a_1,\dots,a_K)$ and any $t$, we have 
\[\reg(\a) - 2\hreg_m(\a)\leq \frac{K(N-\zeta+ D)}{\gamma_m},\]
and 
\[\hreg_m(\a) - 2\reg(\a)\leq \frac{K(N-\zeta+ D)}{\gamma_m},\]
where 
\[\gamma_m = \frac{\sqrt{N-\zeta+ D}}{32\phi_m}\]

Now using this induction hypothesis, we get the following.

\begin{proof}
Again, with our main modifications in place for our reduction to the setting of \citet{sen2021top}, we can rely on their proof for this induction step. See \citet[supplemental p. 5]{sen2021top} for the base case which holds for the $\tau=2$ case. We then seek to show that under the induction hypothesis the bound holds for $\tau \ge 2$.

For simplicity, we omit the dependence on $x$. Introducing extra terms which cancel out, decompose $\reg(\a)$ into three sums: 
\begin{align}
    \reg(\a) &= \E_x \bigg[\sum_{i=1}^K (\pred_\tau(a_i^*) - \pred_\tau(a_i)) + \sum_{i=1}^K (r^*(a_i^*) - \pred_\tau(a_i^*)) + \sum_{i=1}^K (\pred_t(a_i) - r^*(a_i))\bigg]. \label{eq:reg_decompose}
    \numberthis
\end{align}
For the first sum, notice that by definition of $\ha_1,\dots,\ha_K$, we have:
\[\sum_{i=1}^K \pred_t(a_i^*) \leq \sum_{i=1}^K \pred_t(\ha_i).\]
Therefore, 
\begin{equation*}
    \E_x\sum_{i=1}^K (\pred_t(a_i^*) - \pred_t(a_i)) \leq \E_x\sum_{i=1}^K (\pred_t(\ha_i) - \pred_t(x_i)) = \hreg_t(\a). 
\end{equation*}
As \citet{sen2021top} do, for the second sum in Eq. \ref{eq:reg_decompose}, we apply Lemma~\ref{lem:fhat_bound} to obtain:
\begin{align*}
    \sum_{i=1}^K (r^*(a_i^*) - \pred_t(a_i^*)) &\leq \sum_{i=1}^K |r^*(a_i^*) - \pred_t(a_i^*)|  
    \leq K \phi_\tau \sqrt{2(N-\zeta+D)},
\end{align*}

noting that the regret term drops out since the action referenced is the same optimal action.

Then, as \citet{sen2021top} do. We also apply Lemma~\ref{lem:fhat_bound} to the third sum, which gives:
\begin{align*}
    \E_x\sum_{i=1}^K (\pred_t(a_i) - r^*(a_i)) \le \sqrt{2} K\phi_\tau \left((N-\zeta+D) + \gamma_\tau K^{-1}\reg(\a)\right)^{1/2}
\end{align*}

Putting the above together and propagating our ABS dependency on the strata parameters through the rest of the induction proof of \citet{sen2021top}, we get:

\begin{align*}
    \reg(\a) &\leq \hreg_\tau(\a) + \phi_\tau K \sqrt{2(N-\zeta+D)} + \sqrt{2}\phi_\tau K \cdot \left((N-\zeta+D) +\gamma_\tau K^{-1}\reg(\alpha) \right)^{1/2} \\
    &\leq \hreg_\tau(\a) + K[2\phi_\tau  \sqrt{2(N-\zeta+D)} +   \phi_\tau\sqrt{2\gamma_\tau K^{-1}\reg(\a) }]\\
    &\leq \hreg_\tau(\a) + 2K\phi_\tau \sqrt{2(N-\zeta+D)} + K\gamma_\tau \phi_\tau^2 + \frac{1}{2}\reg(\a) 
\end{align*}
and thus
\begin{align*}
    \reg(\a) &\leq 2\hreg_\tau(\a) + 4K\phi_\tau\sqrt{2(N-\zeta+D)} +  2K\gamma_\tau  \phi_\tau^2 \\&\leq 2\hreg_\tau(\a) + \frac{K(N-\zeta+D)}{2\gamma_\tau}.
\end{align*}

Now completing the complementary inequality for $\hreg(\a)$ in an identical fashion.
\begin{align}
    \label{eq:RemplessR}
    \hreg_\tau(\a) &\leq \reg(\a) + \E_{x} \sum_{i=1}^K [\pred_\tau(\ha_i) - r^*(\ha_i)] +  \E_{x} \sum_{i=1}^K [ r^*(a_i)-\pred_\tau(a_i)].
\end{align}

Continuing to follow \citet{sen2021top} and propagate our modified bound, the third sum in Eq. \ref{eq:RemplessR} is bounded by Lemma~\ref{lem:fhat_bound} as follows:
\begin{align*}
    \E_{x} \frac{1}{K}\sum_{i=1}^K | r^*(a_i)-\pred_\tau(x,a_i)| &\leq \sqrt{2} \phi_\tau \left((N-\zeta+D) + \gamma_\tau K^{-1}\reg(\a) \right)^{1/2} \\
    &\leq  \sqrt{2} \phi_\tau \left((N-\zeta+D) + 2\gamma_\tau K^{-1} \hreg_\tau(\a) + \frac{1}{2}(N-\zeta+D) \right)^{1/2} \\
    &\leq 2\phi_\tau \sqrt{(N-\zeta+D)} + 2\phi_\tau^2 \gamma_\tau + \frac{1}{2K}\hreg_\tau(\a) \\
    &\leq \frac{(N-\zeta+D)}{4\gamma_\tau } + \frac{1}{2K}\hreg_\tau(\a).
\end{align*}
The, again following \citet{sen2021top}, we bound the middle term of Eq. \ref{eq:RemplessR}. Note now $\widehat{\a}_\tau = (\ha_{1}, \ldots, \ha_{k})$:
\begin{align*}
    \E_{x} \frac{1}{K}\sum_{i=1}^K [\pred_\tau(\ha_i) - r^*(\ha_i)] \leq \frac{(N-\zeta+D)}{4\gamma_\tau} + \frac{1}{2K} \hreg_\tau(\widehat{\a}_\tau) = \frac{(N-\zeta+D)}{4\gamma_\tau}.
\end{align*}

Putting it all together, we can now bound the model regret:

\begin{align*}
    \hreg_\tau(\a) &\leq 2\reg(\a) + \frac{K(N-\zeta+D)}{\gamma_\tau }.
\end{align*}

And as \citet{sen2021top} point out, $\a$ is arbitrary so the induction step follows.

\end{proof}

\begin{lemma}
    \label{lem:fhat_bound}
    Assume that Lemma~\ref{lem:square_prob_bound} holds and Lemma~\ref{lem:blorg} holds. 
    Recall from Lemma~\ref{lem:blorg} that
    
\begin{align*}
   \E_{x\sim\D} \frac{1}{K}\sum_{i=1}^K \Delta_{i,\tau}         &\leq \phi_\tau \bigg( (N-\zeta + D)  + \frac{1}{t_{\tau-1}}\sum_{t} \gamma_{\hat{\tau}(t)} \E_{x\sim\D}\frac{1}{K} \sum_i (\pred_\tau(x, \hat{a}_i)) - \pred_\tau(x, a_i)\bigg)^{1/2} \\ 
\end{align*}
    
Then assuming the induction hypothesis $(\tau)$ means, if $\gamma_i$ are non-decreasing, that

\begin{align*}
   \E_{x\sim\D} \frac{1}{K}\sum_{i=1}^K \Delta_{i,\tau}         &\leq \sqrt{2}\phi_\tau \bigg( (N-\zeta + D)  + \frac{1}{t_{\tau-1}}\sum_{t} \gamma_{\hat{\tau}(t)} \E_{x\sim\D}\frac{1}{K} \sum_i (r^*(x, a^*_i)) - r^*(x, a_i)\bigg)^{1/2} \\ 
\end{align*}

\end{lemma}

\begin{proof}
See \citet[supplemental p. 5] {sen2021top} for details. We omit the proof for brevity, but note that there is only a minor difference for proving this Lemma, swapping out for the modified $(N-\zeta +D)$ term.
\end{proof}

\begin{proof}[Proof for Theorem~\ref{thm:topkregret}.] Finally, we can leverage \citet[supplemental p. 7-8]{sen2021top} to prove a regret bound incorporating our ABS characteristics. 

Under the same assumptions as \citet{sen2021top}, including that the event in Lemma~\ref{lem:square_prob_bound} holds and Lemma~\ref{lem:fhat_bound} holds, as well as the inductive statements hold. Then we can bound the regret as follows. Let $\ha_1,\dots,\ha_{K-1}$ be greedily chosen arms, and $a_K$ the randomly chosen $K$-th arm. The inductive hypothesis gives 

\begin{align}
      \reg(\ha_1,\dots,\ha_{K-1},a_K) 
     &\leq \frac{K(N-\zeta+ D)}{\gamma_\tau} + 2\E_{x,a_k} \sum_{i=1}^K [\pred_\tau(x,\ha_i)) - \pred_\tau(x,a_i)]  \nonumber \\
    &= \frac{K(N-\zeta+ D)}{\gamma_\tau}+ 2\E_{x,a_k} [\pred_\tau(x,\ha_k)) - \pred_\tau(x,a_k)]  \nonumber \\
    &\le \frac{K(N-\zeta+ D)}{\gamma_\tau}+ 2\E_{x} \sum_{a \notin G} p(a_k | x) [\pred_\tau(x,\ha_k)) - \pred_\tau(x,a_k)] 
\end{align}
For brevity, let's set aside the first term for one moment and focus on the second:
\begin{align}
   2\E_{x} \sum_{a \notin G} p(a_k | x) [\pred_\tau(x,\ha_k)) - \pred_\tau(x,a_k)]  &\le 2\E_{x} \sum_{a \notin G} |S_h|^{-2}\frac{1}{Z} \sum_{a \in S_h} \frac{\pred_\tau(x,\ha_k) - \pred_\tau(x,a_k)}{N-\zeta + \gamma_\tau (\pred_\tau(x,\ha_k)) - \pred_\tau(x,a_k))}  \nonumber
\end{align}

Since $\hat{r}$ is bounded $[0,1]$, like \citet{sen2021top} we note that:

\begin{equation}
    \frac{c}{a + bc} \le \frac{1}{a + b},
\end{equation}

if $c \le 1$ and $a,b,c \ge 0$.

Thus we can bound the regret gap as: 

\begin{align}
   2\E_{x} \sum_{a \notin G} p(a_k | x) [\pred_\tau(x,\ha_k)) - \pred_\tau(x,a_k)] &
   \le 2\E_{x} \sum_{a \notin G} |S_h|^{-2}\frac{1}{Z} \sum_{a \in S_h} \frac{1}{N-\zeta + \gamma_\tau}  \nonumber \\
    &\le 2\E_{x} \sum_{a \notin G} \frac{1}{Z|S_h|} \frac{1}{N-\zeta + \gamma_\tau}.  \nonumber
\end{align}

Now, we find a bound for $\frac{1}{Z}$. Recall that, if $N'=N-\zeta$,  

\begin{equation}
    Z = \sum_{S_h \in S} |S_h|^{-1} \sum_{a \in S_h} \frac{1}{N' + \gamma_\tau (\pred_\tau(x,\ha_k)) - \pred_\tau(x,a_k))}
\end{equation}

Then, since again the maximum gap between rewards is $1$:

\begin{equation}
    Z \ge \sum_{S_h \in S} |S_h|^{-1} \sum_{a \in S_h} \frac{1}{N' + \gamma_\tau }
    \ge \sum_{S_h \in S} \frac{1}{N' + \gamma_\tau }
    \ge \frac{H}{N' + \gamma_\tau }
\end{equation}

And thus:

\begin{equation}
    \frac{1}{Z} \le \frac{N' + \gamma_\tau}{H}
\end{equation}

Returning to the previous bound:

\begin{align}
   2\E_{x} \sum_{a \notin G} p(a_k | x) [\pred_\tau(x,\ha_k)) - \pred_\tau(x,a_k)] 
    &\le 2\E_{x} \sum_{a \notin G} \frac{N - \zeta + \gamma_\tau}{H|S_h|} \frac{1}{N-\zeta + \gamma_\tau}.  \nonumber
        =\le 2\E_{x} \sum_{a \notin G} \frac{1}{H|S_h|}.  \nonumber\\
        \end{align}
Now, recalling that for simplicity we assume that strata are evenly sized and thus $|S_h| = \frac{N-\zeta}{H}$. Thus everything cancels out and we are left with a constant.

\begin{align}
          2\E_{x} \sum_{a \notin G} p(a_k | x) [\pred_\tau(x,\ha_k)) - \pred_\tau(x,a_k)]  &\le 2.  \nonumber
\end{align}

Recall that the event in Lemma~\ref{lem:square_prob_bound} holds with probability at least $1-\delta$ under realizability if we set
\begin{align*}
    \phi_l = \sqrt{\frac{162}{t_{\tau-1}} \log \left( \frac{|\cF| t_{\tau-1}^3}{\delta}\right)} .
\end{align*}

We note that \citet{sen2021top} set $t_{\tau} = 2^\tau \leq 2T$. We note however, that in practice our $t_{\tau} = 2\tau$ due to our reward gap. Thus our true regret bound is strictly smaller by a factor that disappears in the asymptotic. As such, for clarity we keep the step size the same in the bound. Recall that we set $\gamma_\tau = \sqrt{N-\zeta+ D}/(32\phi_\tau)$. We find that the cumulative regret is bounded with probability at least $1-\delta$ by
\begin{align*}
    R(T) &\leq \sum_{\tau=2}^{\hat{\tau}(T)} \frac{K(N-\zeta+ D)t_{\tau-1}}{\gamma_\tau} + 2 \\
    R(T) &\leq \sum_{\tau=2}^{\hat{\tau}(T)} \frac{K(N-\zeta+ D)t_{\tau-1}}{\gamma_\tau} + \sum_{\tau=2}^{\hat{\tau}(T)} 2 \\
    \end{align*}
And recalling that the number of epochs $\tau \le \log_2(2T)$
\begin{align*}
    R(T) &\leq \sum_{\tau=2}^{\hat{\tau}(T)} \frac{K(N-\zeta+ D)t_{\tau-1}}{\gamma_\tau} +  2 \log(2T) \\
    &\leq  2 \log(2T) + \sqrt{162}\cdot32\cdot K\sqrt{(N - \zeta+ D) \log \left( \frac{|\cF| T^3}{\delta}\right)} \sum_{\tau=2}^{\log_2(2T)} 2^{(\tau-1)/2} \\
    & =  \cO \left(\log(T) + K\sqrt{(N-\zeta + D) T  \log \left( \frac{|\cF| T}{\delta} \right)} \right)
\end{align*}

\end{proof}

\section*{Reproducibility Checklist}

\begin{itemize}
    \item Includes a conceptual outline and/or pseudocode description of AI methods introduced. \textbf{Yes, see algorithm box in main text. We also include extended discussion in the appendix and code in the supplemental.}
    \item     Clearly delineates statements that are opinions, hypothesis, and speculation from objective facts and results. \textbf{Yes.}
    \item Provides well marked pedagogical references for less-familiar readers to gain background necessary to replicate the paper. \textbf{Yes.}
    \item Does this paper make theoretical contributions? \textbf{Yes, though they are mainly relegated to the appendix.}
    \item All assumptions and restrictions are stated clearly and formally. \textbf{Yes.}
    \item All novel claims are stated formally (e.g., in theorem statements). \textbf{Yes.}
    \item Proofs of all novel claims are included. \textbf{Yes.}
    \item Proof sketches or intuitions are given for complex and/or novel results. \textbf{Yes.}
 \item Appropriate citations to theoretical tools used are given. \textbf{Yes.}
 \item All theoretical claims are demonstrated empirically to hold. \textbf{Yes.}
    \item All experimental code used to eliminate or disprove claims is included. \textbf{Yes.}
    \item Does this paper rely on one or more datasets? \textbf{Yes.}
    \item     A motivation is given for why the experiments are conducted on the selected datasets. \textbf{Yes.}
    \item All novel datasets introduced in this paper are included in a data appendix. \textbf{No. We are unable to publish even anonymized data due to statutory constraints related to IRS data. 26 U.S. Code \S~6103.}
    \item All novel datasets introduced in this paper will be made publicly available upon publication of the paper with a license that allows free usage for research purposes. \textbf{No. See above.}
  \item  All datasets drawn from the existing literature (potentially including authors’ own previously published work) are accompanied by appropriate citations. \textbf{N/A}
    \item All datasets drawn from the existing literature (potentially including authors’ own previously published work) are publicly available. \textbf{N/A}
    \item All datasets that are not publicly available are described in detail, with explanation why publicly available alternatives are not scientifically satisficing. \textbf{This paper is aimed at the specific IRS application, which is tied to the data that cannot be released by law.}
    \item Does this paper include computational experiments? \textbf{Yes.}
\item Any code required for pre-processing data is included in the appendix. \textbf{Yes.}
\item All source code required for conducting and analyzing the experiments is included in a code appendix. \textbf{Yes.}
\item  All source code required for conducting and analyzing the experiments will be made publicly available upon publication of the paper with a license that allows free usage for research purposes. \textbf{Yes.}
\item All source code implementing new methods have comments detailing the implementation, with references to the paper where each step comes from. \textbf{Yes.}
\item If an algorithm depends on randomness, then the method used for setting seeds is described in a way sufficient to allow replication of results. \textbf{Yes.}
\item This paper specifies the computing infrastructure used for running experiments (hardware and software), including GPU/CPU models; amount of memory; operating system; names and versions of relevant software libraries and frameworks. \textbf{Yes, some information included in code, rest in Appendix.}
\item This paper formally describes evaluation metrics used and explains the motivation for choosing these metrics. \textbf{Yes.}
\item This paper states the number of algorithm runs used to compute each reported result. \textbf{Yes.}
\item Analysis of experiments goes beyond single-dimensional summaries of performance (e.g., average; median) to include measures of variation, confidence, or other distributional information. \textbf{Yes.}
\item The significance of any improvement or decrease in performance is judged using appropriate statistical tests (e.g., Wilcoxon signed-rank). \textbf{We use significance testing in some places, but in most places we give treatment effects and confidence intervals on those effects, which is the new recommended norm.}
\item This paper lists all final (hyper-)parameters used for each model/algorithm in the paper’s experiments. \textbf{Yes.}
\item This paper states the number and range of values tried per (hyper-) parameter during development of the paper, along with the criterion used for selecting the final parameter setting. \textbf{Yes.}

\end{itemize}
    
\fi

\end{document}